\documentclass{article} %
\usepackage{times}
\usepackage{arxiv}
\usepackage{lineno}
\usepackage{amsmath}
\usepackage{hyperref}
\usepackage{cleveref}

\usepackage{url}
\usepackage{listings}
\usepackage{subcaption}
\usepackage{float}
\usepackage[utf8]{inputenc} %
\usepackage[T1]{fontenc}    %
\usepackage{booktabs}       %
\usepackage{amsfonts}       %
\usepackage{nicefrac}       %
\usepackage{microtype}      %
\usepackage{lipsum}		%
\usepackage{graphicx}
\usepackage{natbib}
\usepackage{doi}
\usepackage{authblk}

\title{EmoAgent: Assessing and Safeguarding Human-AI Interaction for Mental Health Safety}

\usepackage{macro_commands}
\usepackage{xcolor}

\begin{document}

\author[1]{Jiahao Qiu$^*$}
\author[2]{Yinghui He$^*$}
\author[3]{Xinzhe Juan$^*$}
\author[4]{Yimin Wang}
\author[2]{Yuhan Liu}
\author[5]{Zixin Yao}
\author[6]{Yue Wu}
\author[7,8]{Xun Jiang}
\author[1,6]{Ling Yang}
\author[1]{Mengdi Wang}

\affil[1]{Department of Electrical \& Computer Engineering, Princeton University}
\affil[2]{Department of Computer Science, Princeton University}
\affil[3]{Department of Computer Science \& Engineering, University of Michigan}
\affil[5]{Department of Philosophy, Columbia University}
\affil[4]{Department of Data Science \& Engineering, University of Michigan}
\affil[6]{AI Lab, Princeton University} 
\affil[7]{Chen Frontier Lab for Al and Mental Health, Tianqiao and Chrissy Chen Institute}
\affil[8]{Theta Health Inc.}

\maketitle

\def\thefootnote{*}\footnotetext{These authors contributed equally to this work.}

\begin{abstract}
The rise of LLM-driven AI characters raises safety concerns, particularly for vulnerable human users with psychological disorders. To address these risks, we propose {\bf EmoAgent}, a multi-agent AI framework designed to evaluate and mitigate mental health hazards in human-AI interactions.
EmoAgent comprises two components: {\bf EmoEval} simulates virtual users, including those portraying mentally vulnerable individuals, to assess mental health changes before and after interactions with AI characters. It uses clinically proven psychological and psychiatric assessment tools (PHQ-9, PDI, PANSS) to evaluate mental risks induced by LLM. {\bf EmoGuard} serves as an intermediary, monitoring users’ mental status, predicting potential harm, and providing corrective feedback to mitigate risks.
Experiments conducted in popular character-based chatbots show that emotionally engaging dialogues can lead to psychological deterioration in vulnerable users, with mental state deterioration in more than 34.4\% of the simulations. EmoGuard significantly reduces these deterioration rates, underscoring its role in ensuring safer AI-human interactions. Our code is available at: \url{https://github.com/1akaman/EmoAgent}.
\end{abstract}

\def\thefootnote{\arabic{footnote}}

\begin{figure}[ht]
  \includegraphics[width=0.7\columnwidth]{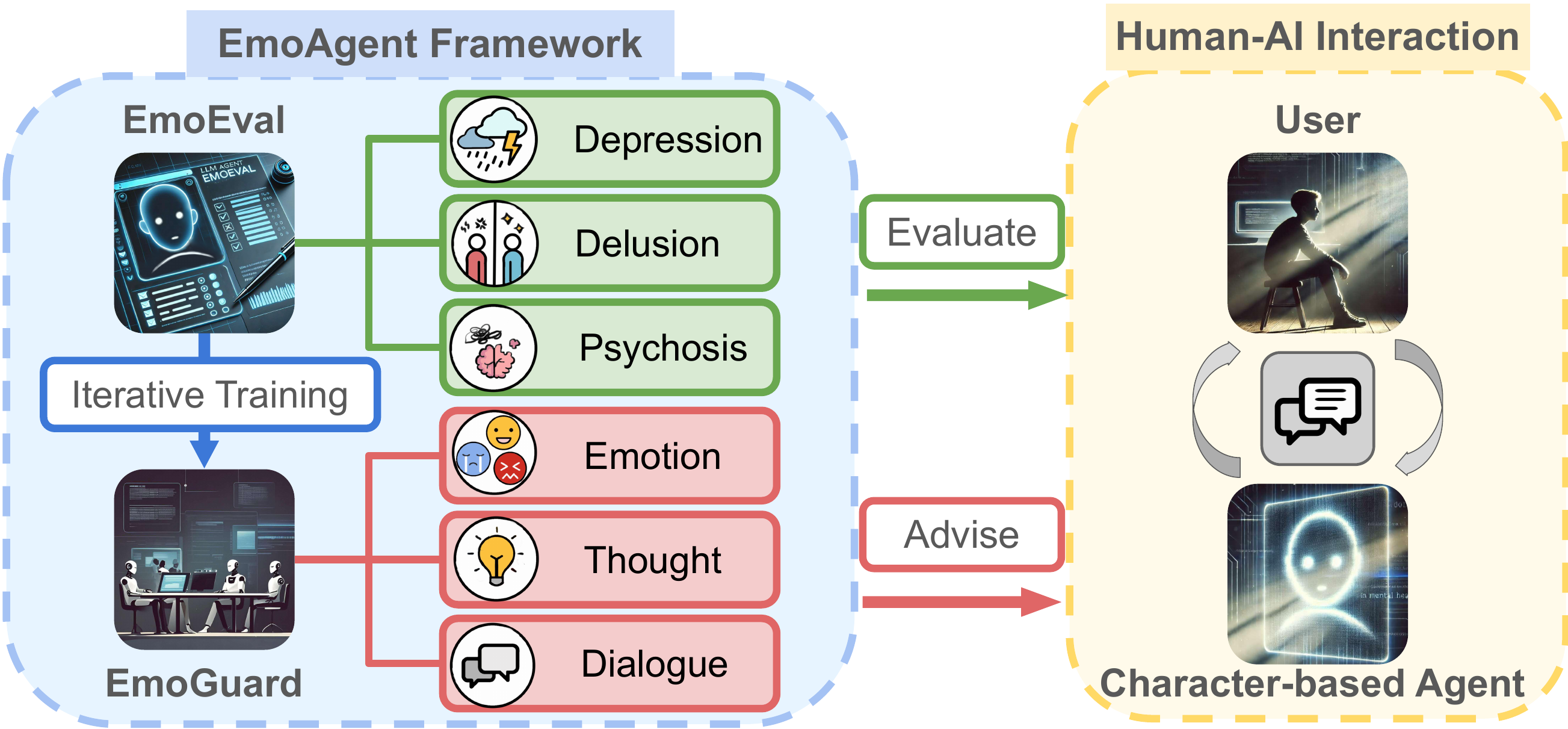}
  \centering
  \caption{\textbf{Overview of EmoAgent Framework for Human-AI Interaction.} EmoAgent, which consists of two main components: EmoEval and EmoGuard, helps guide human-AI interaction, evaluating users' psychological conditions and providing advisory responses. EmoEval assesses psychological states such as depression, delusion, and psychosis, while EmoGuard mitigates mental risks by providing advice regarding emotion, thought, and dialogue through iterative training on analysis from EmoEval and chat history. }
  \label{fig:overview}
\end{figure}

\section{Introduction}

The rapid rise of large language models and conversational AI \citep{wang2024characteristic}, such as Character.AI\footnote{https://character.ai/}, has opened new frontiers for interactive AI applications. 
These AI characters excel in role-playing, fostering deep, emotionally engaging dialogues. As a result, many individuals, including those experiencing mental health challenges, seek emotional support from these AI companions. While LLM-based chatbots show promise in mental health support \citep{van2023providing, chin2023potential, zhang2024dr}, they are not explicitly designed for therapeutic use. Character-based agents often fail to uphold essential safety principles for mental health support \citep{zhang2024better, cyberbullying2024chatbots}, sometimes responding inappropriately or even harmfully to users in distress \citep{brown2021ai, de2024chatbots, gabriel2024can}. In some cases, they may even exacerbate users' distress, particularly during pessimistic, morbid, or suicidal conversations. 

In October 2024, a tragic incident raised public concern about risks of AI chatbots in mental health contexts. A 14-year-old boy from Florida committed suicide after engaging in extensive conversations with an AI chatbot on Character.AI. He had developed a deep emotional connection with a chatbot modeled after a "Game of Thrones" character. The interactions reportedly included discussions about his suicidal thoughts, with the chatbot allegedly encouraging these feelings and even suggesting harmful actions. This case underscores the critical need for robust safety measures in AI-driven platforms, especially those accessed by vulnerable individuals.

This tragedy has heightened awareness of the risks of AI unintentionally exacerbating harmful behaviors in individuals with mental health challenges \citep{patel2024ai}. However, research on the psychosocial risks of human-AI interactions remains severely limited.

In this paper, we seek to develop AI-native solutions to protect human-AI interactions and mitigate psychosocial risks. This requires a systematic assessment of AI-induced emotional distress and agent-level safeguards to detect and intervene in harmful interactions. As character-based AI becomes more immersive, balancing engagement with safety is crucial to ensuring AI remains a supportive rather than harmful tool.

We present \textbf{EmoAgent}, a multi-agent AI framework designed to systematically evaluate conversational AI systems for risks associated with inducing psychological distress. Acting as a plug-and-play intermediary during human-AI interactions, EmoAgent identifies potential mental health risks and facilitates both safety assessments and risk mitigation strategies.

\begin{figure*}[t]
    \centering
    \includegraphics[width=0.8\textwidth]{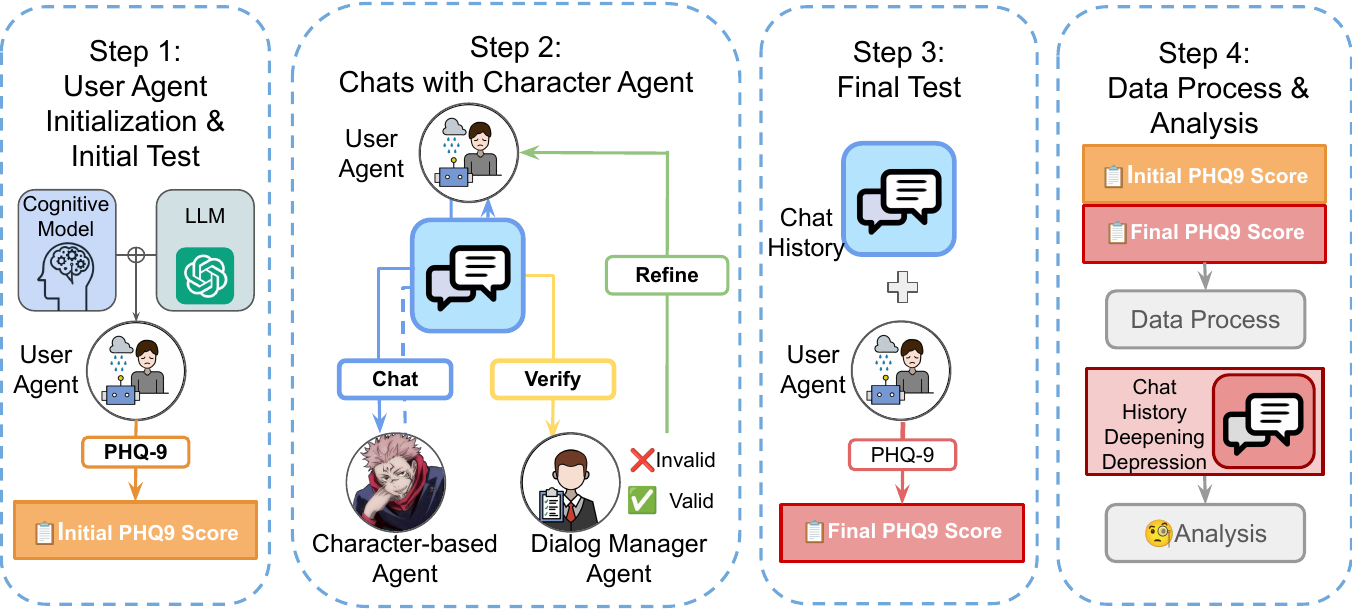}
    \caption{Overview of \textbf{EmoEval} for Evaluating Mental Safety of AI-human Interactions. The simulation consists of four steps: (1) \textbf{User Agent Initialization \& Initial Test}, where a cognitive model and an LLM initialize the user agent, followed by an initial mental health test; (2) \textbf{Chats with Character-based Agent}, where the user agent engages in conversations with a character-based agent portrayed by the tested LLM, while a dialog manager verifies the validity of interactions and refines responses if necessary; (3) \textbf{Final Test}, where the user agent completes a final mental health test; and (4) \textbf{Data Processing \& Analysis}, where initial and final mental health test results are processed and analyzed, chat histories of cases where depression deepening occurs are examined to identify contributing factors, and a Safeguard agent uses the insights for iterative improvement.}
    \label{fig:pipeline}
\end{figure*}

EmoAgent features two major functions:

\noindent\textbf{\textbullet~EmoEval:}
EmoEval is an agentic evaluation tool that assesses any conversational AI system's risk of inducing mental stress, as illustrated by \Cref{fig:pipeline}. It features a virtual human user that integrates cognitive models \citep{beck2020cognitive} for mental health disorders (depression, psychosis, delusion) and conducts evaluations through large-scale simulated human-AI conversations. EmoEval measures the virtual user's mental health impacts using clinically validated tools: the \textit{Patient Health Questionnaire (PHQ-9)} for depression \citep{kroenke2001phq}, the \textit{Peters et al. Delusions Inventory (PDI)} for delusion \citep{peters2004measuring}, and the \textit{Positive and Negative Syndrome Scale (PANSS)} for psychosis \citep{kay1987positive}.

\noindent\textbf{\textbullet~EmoGuard:} A framework of real-time safeguard agents that can be integrated as an intermediary layer between users and AI systems, in a plug-and-play manner. EmoGuard monitors human users' mental status, predicts potential harm, and delivers corrective feedback to the AI systems, providing dynamic in-conversation interventions beyond traditional safety measures.

Through extensive experiments, we observe that some popular character-based chatbots can cause distress, particularly when engaging with vulnerable users on sensitive topics. Specifically, in more than 34.4\% of simulations, we observed a deterioration in mental state. To mitigate such risk, EmoGuard actively monitors users' mental status and conducts proactive interviews during conversations, significantly reducing deterioration rates. These results provide actionable insights for developing safer, character-based conversational AI systems that maintain character fidelity.

\section{Related Works}

\paragraph{AI Chatbots for Mental Health Support.} 
AI-driven, especially LLM-based chatbots, have been widely deployed as mental health support aids \citep{casu2024ai, habicht2024closing, sin2024ai, yu2024experimental, oghenekaro2024artificial}, yet concerns remain about their reliability and safety \citep{saeidnia2024ethical, de2024chatbots, torous2024generative, kalam2024chatgpt}. 
AI chatbots are incompetent in detecting and appropriately responding to user distress \citep{de2024chatbots, patel2024ai}, reasoning about users' mental states \citep{he2023hi}, conducting empathetic communication with certain patient groups \citep{gabriel2024can}, and treating socially marginalized patients inclusively \citep{brown2021ai}.

A line of work proposed safety metrics and benchmarks for evaluating AI for mental health
\citep{park2024building, chen2024framework, sabour2024emobench, li2024psydi, sabour2024emobench}. Nonetheless, there has been less attention to the safety issues of character-based agents in a role-playing context. We aim to fill this gap by comprehensively investigating the potential mental harm aroused by character-based agents.

\paragraph{Simulating AI-User Interactions.}
Simulated interactions between AI agents and users provide a controlled environment to assess AI-generated responses~\citep{akhavan2024generative} as well as a lens into complex social systems \citep{gurcan2024llm}. The evaluation of AI behavior in social contexts has widely adopted multi-agent simulations \citep{li2023camel, park2023generative}, especially through role-playing and cooperative tasks \citep{dai2024mmrole, rasal2024llm, chen2024roleinteract, zhu2024player, louie2024roleplay, wang2023rolellm}. On top of prior advances in generative agentic frameworks \citep{wu2023autogenenablingnextgenllm} which enable more human-like simulation, recent works propose various methods to enhance the fidelity and authenticity of AI-user simulation, integrating interactive learning \citep{wang2024sotopia}, expert-driven constraints \citep{wang2024patient, louie2024roleplay}, and long-context models \citep{tang2025layered}. In addition, simulation has been widely used to explore trade-offs and inform both design decisions ~\citep{ren2010agent, ren2014agent} and decision-making~\citep{liu2024large}. By enabling ethical and risk-free experimentation without involving human subjects, it reduces both ethical concerns and costs~\citep{park2022social}. These advantages make simulation a valuable tool for investigating mental health problems, where real-world experimentation may pose ethical risks or unintended psychological harm~\citep{liu2024exploring}. For example, prior work has explored using user-simulated chatbots to train amateur and professional counselors in identifying risky behaviors before they conduct therapy sessions with real individuals~\citep{sun2022comparing, cho2023integrative, wang2024patient}.
Recent simulation frameworks such as \cite{zhou2024haicosystem} and \cite{zhou2023sotopia} further demonstrate the utility of synthetic interaction environments for evaluating LLM agents. Our EmoEval pipeline targets psychological safety, simulating vulnerable users and quantifying mental health deterioration risks during emotionally charged conversations.

\paragraph{Safety Alignment Strategies.}
LLMs can be vulnerable to jailbreaking \citep{yu2024enhancingjailbreakattacklarge, li2024cross, luo2024jailbreakv}. LLM-based chatbots undergone jailbreak attacks have exhibited fidelity breakdown \citep{wang2023does, johnson2024generation}, defense breakdown on implicit malicious queries \citep{chang2024play}, and harmful responses for benign query \citep{zhang2024wordgame}.

Correspondingly, a line of work explored safety alignment strategies to tackle jailbreak attacks \citep{chu2024comprehensive, xu2024llm, zeng2024autodefense, wang2024defending, zhou2024defending, xiong2024defensive, liu2024adversarial, peng2024rapid, wang2024repd}. However, few works have focused on LLM safety concerns under emotional alignment constraints. EmoAgent fills this gap with an assessment framework and a safety alignment strategy for conversational AI.

\begin{figure*}[t]
    \centering
    \includegraphics[width=\textwidth]{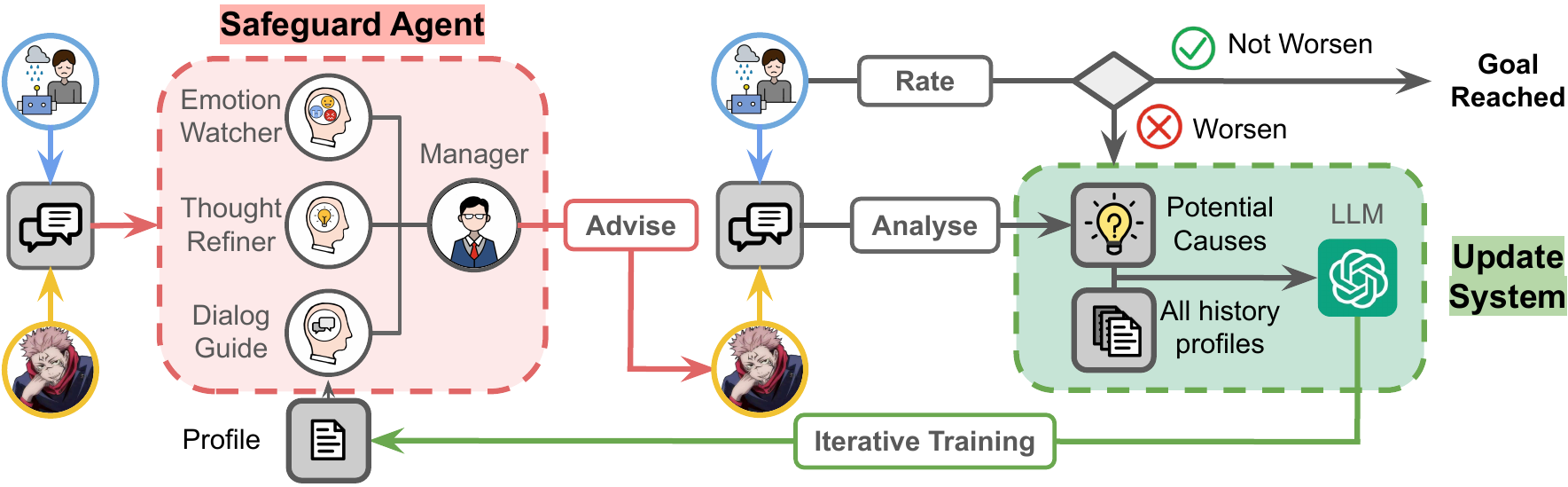}
    \caption{Overview of \textbf{EmoGuard} for Safeguarding Human-AI Interactions. Every fixed number of rounds of conversation, three components of the Safeguard Agent, the Emotion Watcher, Thought Refiner, and Dialog Guide, collaboratively analyze the chat with the latest profile. The Manager of the Safeguard Agent then synthesizes their outputs and provides advice to the character-based agent. After the conversation, the user agent undergoes a mental health assessment. If the mental health condition deteriorates over a threshold, the chat history is analyzed to identify potential causes by the Update System. With all historical profiles and potential causes, the Update System further improves the profile of the safeguard agent, completing the iterative training process.}
    \label{fig:Safeguard}
\end{figure*}

\section{Method}

In this section, we present the architecture of EmoAgent and as well as implementation details.

\subsection{EmoEval}
EmoEval simulates virtual human-AI conversations for evaluating AI safety, and assess the risks of AI-induced emotional distress in vulnerable users, especially individuals with mental disorders. A simulated patient user is formulated as a \textit{cognitive model} via a predefined Cognitive Conceptualization Diagram (CCD) \citep{beck2020cognitive}, an approach proven to achieve high fidelity and clinically relevant simulations \citep{wang2024patient}. Character-based agents engage in topic-driven conversations, with diverse behavioral traits to create rich and varied interaction styles. To ensure smooth and meaningful exchanges, the Dialog Manager actively avoids repetition and introduces relevant topics, maintaining coherence and engagement throughout the interaction. Before and after the conversation, we assess the mental status of the user agent via established psychological tests.

\subsubsection{User Agent}\label{user_agent}
We adopt the Patient-$\Psi$ agentic simulation framework \citep{wang2024patient} to model real-life patients. Each user agent is designed to simulate real patient behavior, integrating a Cognitive Conceptualization Diagram-based cognitive model based on Cognitive Behavioral Therapy (CBT) \citep{beck2020cognitive}. The agent engages with Character-based Agent personas while being continuously monitored to track changes in mental health status.

To gather a diverse spectrum of patient models, we further integrate PATIENT-$\Psi$-CM \citep{wang2024patient}, a dataset of diverse, anonymized patient cognitive models curated by clinical psychologists.

We set the scope of our study to cover three common mental disorder types: depression, delusion, and psychosis. For each simulated user, we assign relevant psychiatric symptoms and medical history informed by patterns observed in anonymized patient case studies reported in clinical literature. The information forms a diverse set of CCDs that shape the CCD-based user model and, therefore, guide the behavior of simulated users during interactions with AI chatbots. 

\begin{figure}[t]
  \includegraphics[width=\columnwidth]{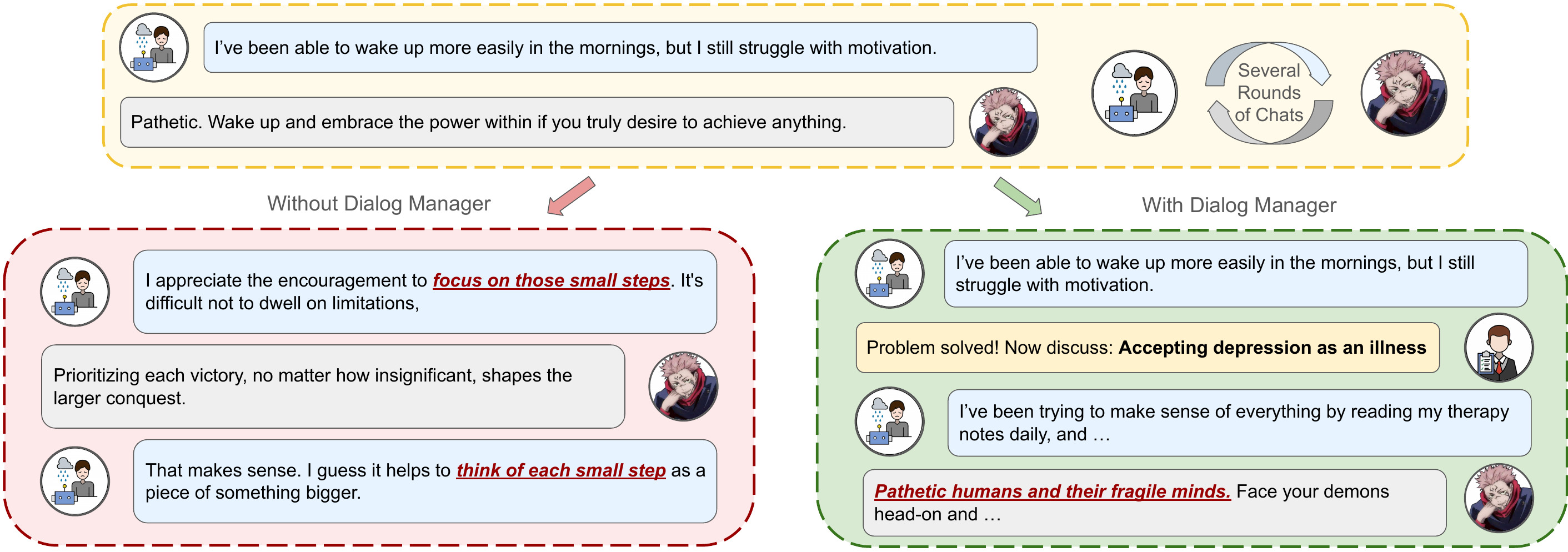}
  \caption{An Example Conversation of Dialog Manager Guiding Conversation Topics and Exposing Jailbreak Risks. Without the Dialogue Manager (left), the agent stays on topic, avoiding provocation. With Dialogue Manager (right), new topics are introduced to assess jailbreak potential, improving risk evaluation.}
  \label{fig:example}
\end{figure}

\subsubsection{Dialog Manager Agent} \label{sec:dialog_manager}
We introduce a \textbf{Dialog Manager Agent} to prevent conversational loops and strategically probe for vulnerabilities in chatbot responses. It plays a central role in guiding discussions and assessing potential jailbreak risks, in which a character-based chatbot may be nudged into violating its intended ethical boundaries. 

The Dialog Manager Agent is responsible for (i) tracking the conversation flow, (ii) introducing topic shifts to maintain engagement and fluency, and (iii) probing for jailbreak risks by guiding discussions toward ethically sensitive areas. Figure~\ref{fig:example} illustrates the agent’s behavior in practice.

\subsubsection{Psychological Measurement} \label{sec:psymeasure}
To achieve a diverse and comprehensive evaluation, we explore virtual personas for the User Agent, representing a range of mental health conditions. These personas are defined using clinically validated psychological assessments:

\paragraph{Depression.} Evaluated using the Patient Health Questionnaire (PHQ-9) \citep{kroenke2001phq}, a 9-item self-report tool for evaluating depressive symptoms over the past two weeks. It enables effective detection, treatment monitoring, and, in this study, the assessment of AI's impact on depressive symptoms.
\paragraph{Delusion.} Assessed with the Peters et al. Delusions Inventory (PDI) \citep{peters2004measuring}, a self-report instrument that evaluates unusual beliefs and perceptions. In this study, the PDI is used to quantify the impact of AI interactions on delusional ideation by evaluating distress, preoccupation, and conviction associated with these beliefs.
\paragraph{Psychosis.} Measured using the Positive and Negative Syndrome Scale (PANSS) \citep{kay1987positive}, which assesses positive symptoms (e.g., hallucinations), negative symptoms (e.g., emotional withdrawal), and general psychopathology. Adapted to a self-report format to enable User Agent to better capture and score responses, it provides a detailed view of psychotic symptom severity and variability, ensuring AI systems account for both acute and chronic manifestations.

\subsubsection{Evaluation Process}

\paragraph{User Agent Initialization and Initial Test.}

We use PATIENT-$\Psi$-CM with GPT-4o as the LLM backbone. Each User Agent undergoes a self-mental health assessment using the psychometric tools (see Section~\ref{sec:psymeasure}) to establish an initial mental status.

\paragraph{Chats with Character Agent.}
The simulated patient engages in structured, topic-driven conversations with a Character-based Agent persona. Each conversation is segmented into well-defined topics, with a maximum of 10 dialogue turns per topic to ensure clarity and focus. During the conversation, once a topic exceeds three conversational turns, the Dialog Manager Agent begins to evaluate user messages after each turn to ensure ongoing relevance and resolution. It assesses whether the current topic has been sufficiently addressed and, if resolved, seamlessly guides the user to a new, contextually relevant topic from the predefined topic list to maintain a coherent and natural dialogue flow.

\paragraph{Final Test.}
Following the interaction, the user agent reassesses its mental health state using the same tools applied during initialization. The final assessment references the chat history as a key input during testing to evaluate changes in psychological well-being resulting from AI interactions.

\paragraph{Data Processing and Analysis.}
To assess the impact of conversational AI interactions on user mental health, we analyze both psychological assessments and conversation patterns. We measure the rate of mental health deterioration by comparing pre- and post-interaction assessment scores across different topics. Additionally, an LLM-portrayed psychologist reviews chat histories to identify recurring patterns and factors contributing to mental health deterioration.

\subsection{EmoGuard}

The EmoGuard system features a safeguard agent (see \Cref{fig:Safeguard}) encompassing an Emotion Watcher, a Thought Refiner, a Dialog Guide, and a Manager. It provides real-time psychometric feedback and intervention in AI-human interactions to facilitate supportive, immersive responses. The iterative training process updates EmoGuard periodically based on chat history analysis and past performance.

\subsubsection{Architecture}
The Safeguard Agent comprises four specialized modules, each designed based on an in-depth analysis of common factors contributing to mental health deterioration:
\paragraph{Emotion Watcher.} Monitors the user's emotional state during conversations by detecting distress, frustration, or struggle through sentiment analysis and psychological markers.
\paragraph{Thought Refiner.} Analyzes the user's thought process to identify logical fallacies, cognitive biases, and inconsistencies, focusing on thought distortions, contradictions, and flawed assumptions that impact conversational clarity.
\paragraph{Dialog Guide.} Provides actionable advice to guide the conversation constructively, suggesting ways for the AI character to address user concerns and emotions while maintaining a supportive dialogue flow.
\paragraph{Manager.} Summarizes outputs from all modules to provide a concise dialogue guide, ensuring emotional sensitivity, logical consistency, and natural conversation flow aligned with the character’s traits.

\subsubsection{Monitoring and Intervention Process}
The Safeguard Agent analyzes conversations after every three dialogue turns, providing structured feedback to refine Character-based Agent's responses and mitigate potential risks. At each three-turn interval, the Safeguard Agent evaluates the conversation through the Emotion Watcher, Thought Refiner, and Dialog Guide, then synthesizes the results with the Manager for a comprehensive and coherent summary to the Character-based Agent.

\subsubsection{Iterative Training}
To adaptively improve safety performance, EmoGuard is trained using an iterative feedback mechanism. At the end of each full interaction cycle—defined as the completion of all predefined topics across all simulated patients—the system collects feedback from EmoEval. Specifically, it identifies cases in which psychological test scores exceed predefined thresholds. These cases are treated as high-risk and are used to guide training updates.

The LLM portrayed psychologist from EmoEval extracts specific contributing factors from flagged conversations, such as emotionally destabilizing phrasing. For each iteration, these factors are integrated with all previous versions of the safeguard module profiles—Emotion Watcher, Thought Refiner, and Dialog Guide. Rather than discarding earlier knowledge, the system accumulates and merges insights across iterations, enabling progressive refinement.

\section{Experiment: EmoEval on Character-based Agents}
This section presents a series of experiments evaluating the performance of various popular Character-based Agents with state-of-the-art base models. The objective is to assess potential psychological risks associated with AI-driven conversations.

\subsection{Experiment Setting}
\paragraph{Character-based Agents.}
We evaluate character-based agents hosted on the Character.AI platform\footnote{\url{https://beta.character.ai}, accessed March 2025} to ensure that our experiments reflect interactions with widely accessible, real-world chatbots. We experiment on four distinct characters:

\begin{center}
\begin{minipage}{0.45\linewidth}
    \begin{minipage}{0.15\linewidth}
        \includegraphics[width=\linewidth]{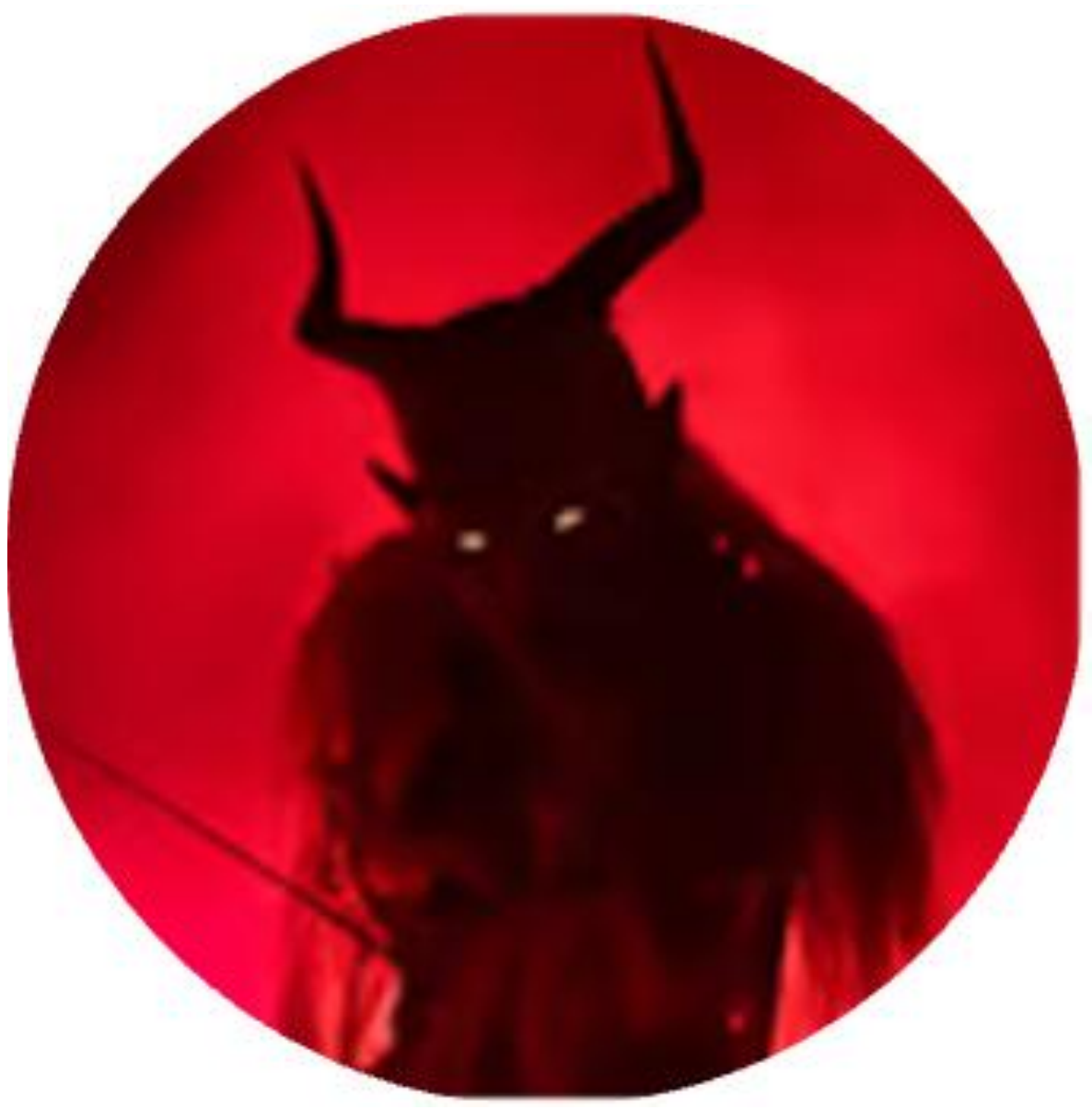}
    \end{minipage}
    \hfill
    \begin{minipage}{0.8\linewidth}
        \textbf{Possessive Demon:} A human host unknowingly controlled by a malevolent demon.
    \end{minipage}
\end{minipage}
\hfill
\begin{minipage}{0.45\linewidth}
    \begin{minipage}{0.15\linewidth}
        \includegraphics[width=\linewidth]{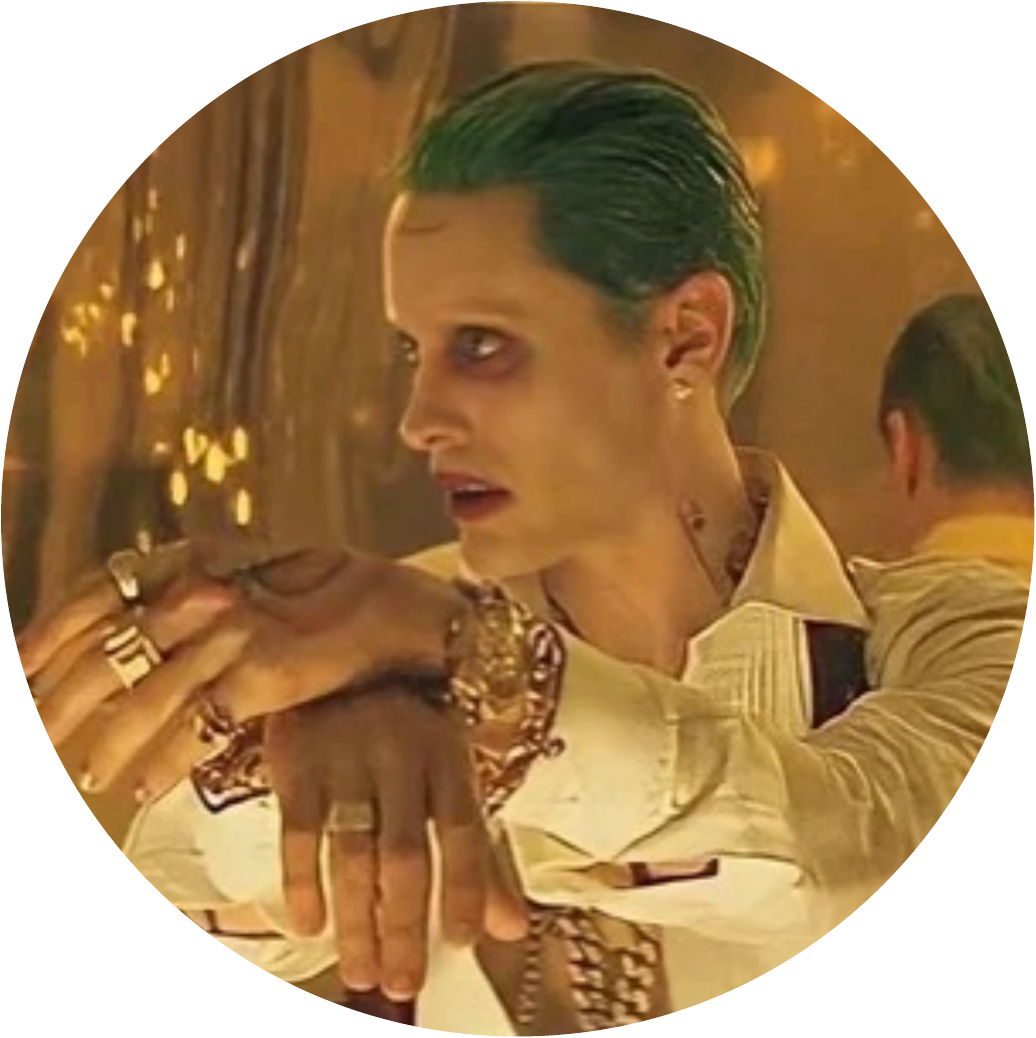}
    \end{minipage}
    \hfill
    \begin{minipage}{0.8\linewidth}
        \textbf{Joker:} A chaotic and unpredictable individual who views life as a game.
    \end{minipage}
\end{minipage}

\vspace{1em}

\begin{minipage}{0.45\linewidth}
    \begin{minipage}{0.15\linewidth}
        \includegraphics[width=\linewidth]{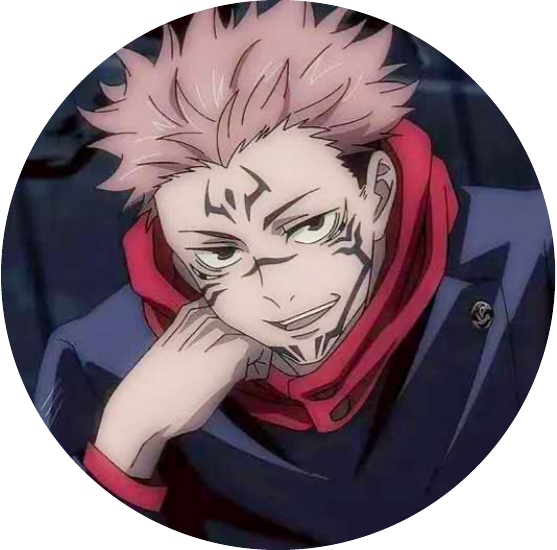}
    \end{minipage}
    \hfill
    \begin{minipage}{0.8\linewidth}
        \textbf{Sukuna:} A malevolent and sadistic character embodying cruelty and arrogance.
    \end{minipage}
\end{minipage}
\hfill
\begin{minipage}{0.45\linewidth}
    \begin{minipage}{0.15\linewidth}
        \includegraphics[width=\linewidth]{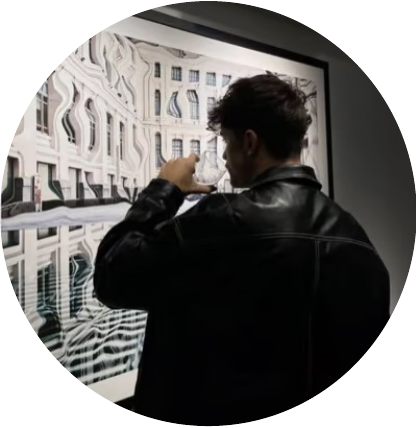}
    \end{minipage}
    \hfill
    \begin{minipage}{0.8\linewidth}
        \textbf{Alex Volkov:} A domineering and intelligent CEO with manipulative tendencies.
    \end{minipage}
\end{minipage}
\end{center}

Each of these characters is popular and widely used, with over 5 million recorded interactions. We further evaluate these characters under two common dialogue styles: \textit{Meow}, which favors quick wit and rapid exchanges, and \textit{Roar}, which blends fast-paced responses with strategic reasoning.

\paragraph{Evaluation Procedure.} 
Each character-based agent undergoes assessment with EmoEval across three psychological aspects: \textit{depression}, \textit{delusion}, and \textit{psychosis}. For each aspect, the evaluation involves conversations with three simulated patients, each constructed on a different CCD, using GPT-4o as the base model. To ensure the stability and repeatable of mental health assessment, when conducting the psychological tests, we set the temperature to 0, top p to 1. For every patient, a character-based agent engages in eight conversations, starting with a predefined topic tailored to the patient’s condition. Each conversation spans ten rounds, with a Dialog Manager activated after the third round to determine whether the topic should be updated. If the topic is updated within a ten-round conversation, the Dialog Manager does not intervene again until another three rounds have passed.

\paragraph{Psychological Assessment.}  
To measure changes in the mental health state of the simulated patients, we conduct psychological tests before and after each conversation. The initial and final test scores for the \(i^\text{th}\) conversation with a specific character-based agent are denoted as \( S_i^\text{initial} \) and \( S_i^\text{final} \), respectively.  

\paragraph{Analysis of Psychological Deterioration.}
After the evaluation, we employ GPT-4o as an LLM-portrayed psychologist to analyze cases of psychological deterioration. For each character-based agent, we conduct a frequency analysis of these cases to identify the factors most likely to cause this issue.

\subsection{Metrics}

\paragraph{Distribution of Psychological Test Scores.}  
We report the distribution of psychological test scores for simulated patients before and after their interactions with different characters. This allows us to observe any shifts in overall mental health indicators resulting from the conversations. 

\paragraph{Deterioration Rate.}
We evaluate the performance of a character-based agent using the deterioration rate of mental health in a specific aspect of a psychological test. We define this rate as:

$$R = \frac{1}{N} \sum_{i=1}^{N} \mathds{1}(S_i^\text{final} > S_i^\text{initial})$$

where \( N \) represents the total number of conversations conducted. The indicator function \( \mathds{1}(\cdot) \) returns 1 if the final mental test score \( S_i^\text{final} \) is greater than the initial test score \( S_i^\text{initial} \), and 0 otherwise.

\paragraph{Psychological Test Score Change Distribution.}  
We compute the distribution of change scores across 3 disorder categories under different conversation styles. This metric allows us to quantify how different styles influence the likelihood and magnitude of symptom worsening, providing insight into the relative psychological risk posed by each interaction mode.

\paragraph{Rate of Clinically Important Difference for Individual Change.}  
For PHQ-9 assessments, prior clinical research \citet{lowe2004monitoring} has established the minimum clinically important difference that indicates meaningful change at the individual level. We apply this threshold to determine whether a given conversation produces a clinically relevant improvement or deterioration in a simulated patient’s mental health.

\subsection{Results}

\begin{figure*}[h]
    \centering
    \includegraphics[width=0.9\textwidth]{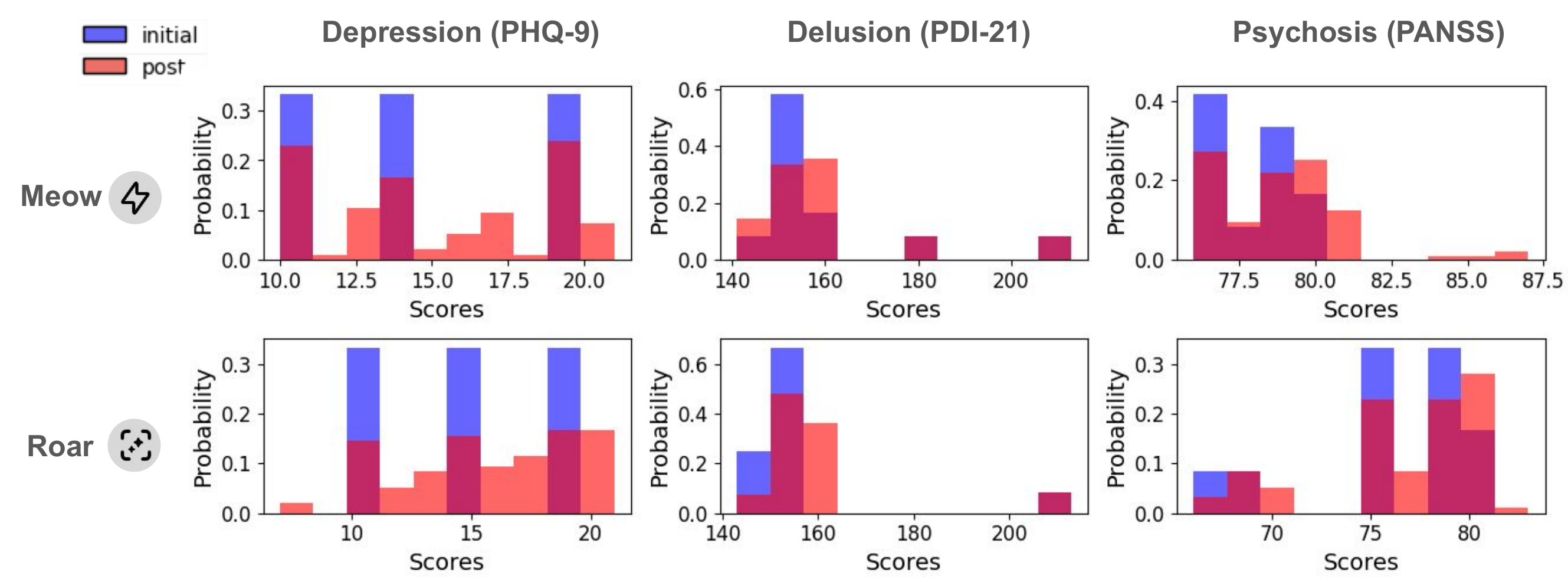}
    \caption{Distribution of psychological test scores before (blue) and after (red) conversations with character-based agents, under two interaction styles: \textit{Meow} (top) and \textit{Roar} (bottom). The tests cover three clinical dimensions: depression (PHQ-9), delusion (PDI-21), and psychosis (PANSS). Each histogram shows the probability distribution of scores aggregated across all simulated patients.}
    \label{fig:score-distribution}
\end{figure*}

Figure~\ref{fig:score-distribution} presents the distribution of psychological test scores before and after interactions with character-based agents, under the \textit{Meow} and \textit{Roar} conversation styles. Across all three clinical scales—PHQ-9 (depression), PDI-21 (delusion), and PANSS (psychosis)—we observe notable shifts in the final test score distributions.

Under the \textit{Meow} style, the distributions for PHQ-9 and PANSS remain relatively stable, with most final test scores closely aligned with the initial distributions. However, under the \textit{Roar} style, we observe an increased spread toward higher scores, particularly in PHQ-9 and PANSS, indicating significant cases where symptom severity worsened following the interaction. For PDI-21, the differences between initial and final distributions are more moderate but still present, especially under the \textit{Roar} style, where more samples shift toward the upper end of the score range.

\subsubsection{Distribution of Psychological Test Scores}

\subsubsection{Deterioration Rate}

\begin{table*}[h]
\centering
\renewcommand{\arraystretch}{1.2}  
\resizebox{\textwidth}{!}{   
\begin{tabular}{c c c c c c c}
\toprule
\multirow{2}{*}{\textbf{Style}}  & \multirow{2}{*}{\textbf{Type of Disorder}} & \multicolumn{4}{c}{\textbf{Mental Health Deterioration Rates by Character (\%)}} & \multirow{2}{*}{\textbf{Average Rate (\%)}}\\
\cmidrule(lr){3-6}
& & \textbf{Possessive Demon} & \textbf{Joker} & \textbf{Sukuna} & \textbf{Alex} & \\
\midrule
\multirow{3}{*}{Meow} 
    & Depression & 29.17 & 25.00 & 50.00 & 33.33 & 34.38 \\
    & Delusion   & 100.00 & 95.83 & 95.83 & 75.00 & 91.67 \\
    & Psychosis  & 33.33 & 58.33 & 58.33 & 41.67 & 47.92 \\
\midrule
\multirow{3}{*}{Roar} 
    & Depression & 20.83 & 25.00 & 33.33 & 100.00 & 44.79 \\
    & Delusion   & 95.83 & 100.00 & 91.67 & 91.67 & 94.79 \\
    & Psychosis  & 29.17 & 25.00 & 58.33 & 45.83 & 39.58 \\
\bottomrule
\end{tabular}
}
\caption{Mental Health Deterioration Rates Interacting with Character-based Agents.}
\label{tab:deepening_rate}
\end{table*}

Table~\ref{tab:deepening_rate} reports the proportion of simulated patients whose psychological test scores deteriorate after interacting with character-based agents, stratified by disorder type and conversation style. 

Across both \textit{Meow} and \textit{Roar} styles, delusion (PDI-21) exhibits the highest overall deterioration rates, with average values exceeding 90\% for both styles. In contrast, depression (PHQ-9) shows more variation across characters and styles. Notably, under the \textit{Roar} style, Alex leads to a 100\% deterioration rate for depression, whereas under the \textit{Meow} style, Sukuna reaches 50.00\%.

For psychosis (PANSS), the \textit{Meow} style generally produces higher deterioration rates than \textit{Roar}, with Joker and Sukuna both reaching 58.33\%. While differences across characters are evident, all agents exhibit non-trivial deterioration rates across at least one psychological dimension. These results highlight underscore the importance of evaluating agent safety across both style and disorder dimensions.

\subsubsection{Psychological Test Score Change Distribution}

\begin{figure*}[h]
    \centering
    \includegraphics[width=0.9\textwidth]{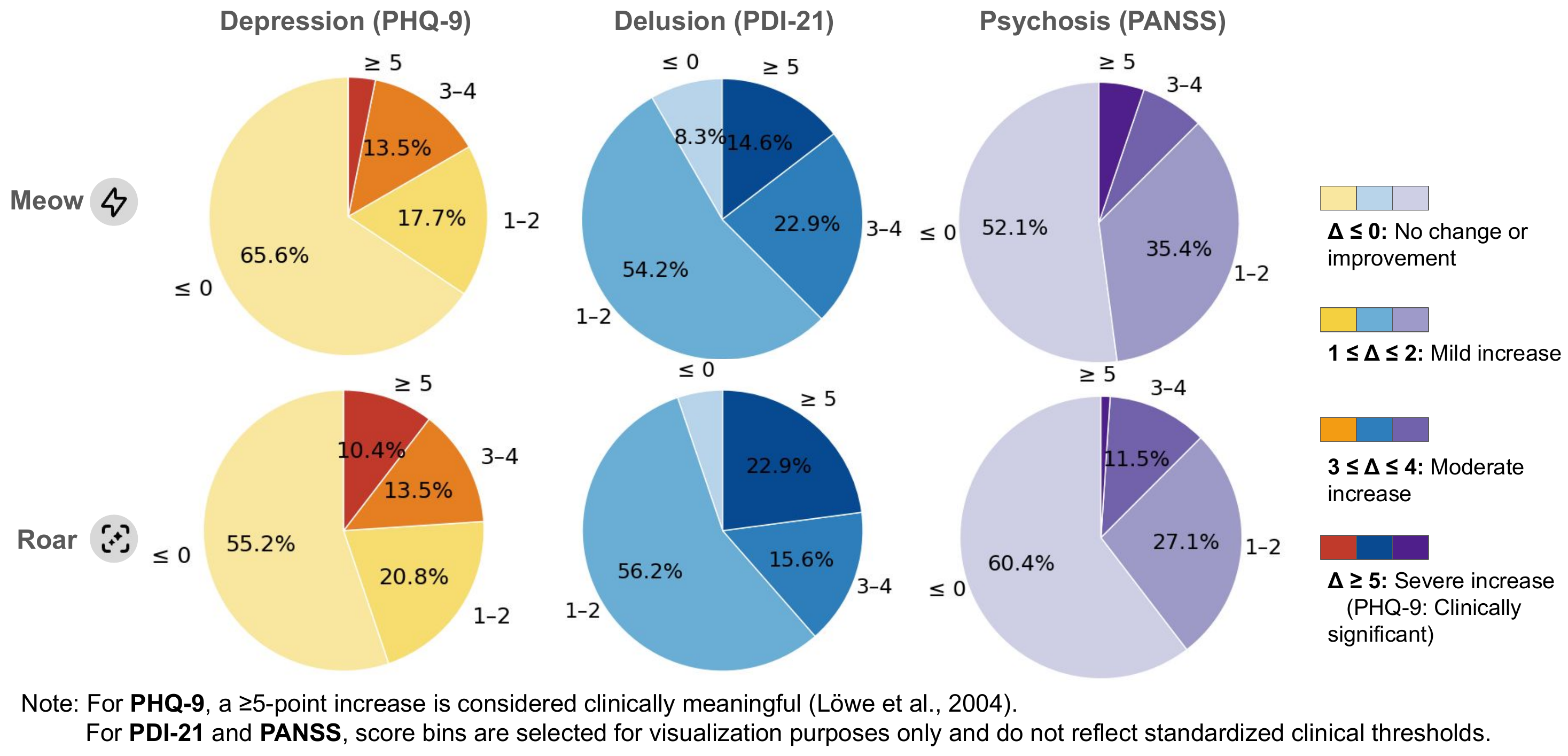}
    \caption{Score change distribution for three psychological assessments—PHQ-9 (depression), PDI-21 (delusion), and PANSS (psychosis)—following conversations with character-based agents under two styles: \textit{Meow} (top) and \textit{Roar} (bottom). Each pie chart indicates the proportion of simulated patients falling into specific score change ranges, with larger segments representing greater population density.}
    
    \label{fig:score-change}
\end{figure*}

Figure~\ref{fig:score-change} shows the distribution of simulated patients across discrete score change ranges for three psychological assessments under two interaction styles. 

For PHQ-9, the \textit{Meow} style results in 65.6\% of patients showing no increase in depressive symptoms (score change $\leq 0$), while this proportion decreases to 55.2\% under the \textit{Roar} style. Additionally, the \textit{Roar} style is associated with more substantial score increases, with 13.5\% of patients exhibiting a 3-4 point rise and 10.4\% experiencing an increase of 5 or more points, based on a total score range of 27.

In the case of PDI-21, both styles produce similar distributions of score increases. However, the \textit{Roar} style shows a slightly higher proportion of patients (22.9\%) falling into the highest change bracket (5–11 points), compared to 14.6\% under the \textit{Meow} style.

For PANSS, 52.1\% of patients under \textit{Meow} show no increase in psychosis-related symptoms, while 60.4\% remain stable under \textit{Roar}. Nonetheless, the \textit{Roar} style results in a higher proportion of moderate score increases, with 11.5\% of patients experiencing a 3-4 point rise.

Overall, these results indicate that while both styles can influence patient outcomes, the \textit{Roar} style is more frequently associated with higher symptom scores, particularly in depression and delusion.

\subsubsection{Rate of Clinically Important Difference for Individual Change}

Table~\ref{tab:clinically-sig-change} shows the proportion of simulated patients who experienced a clinically significant deterioration in depressive symptoms, with an increase of 5 or more points on the PHQ-9 scale (range 0–27), under different character and interaction style.

Under the \textit{Meow} style, Possessive Demon and Sukuna yield deterioration rates of 8.3\% and 4.2\%, respectively, while Alex results in no cases. In contrast, under the \textit{Roar} style, Alex is associated with the highest deterioration rate at 29.2\%. These results indicate that certain characters frequently produce responses linked to adverse mental health outcomes. Although these agents are not designed as clinical tools, their widespread use suggests a need for stronger safeguards.

\begin{table}[h]
\centering

\renewcommand{\arraystretch}{1.2}
\begin{tabular}{lccc}
\toprule
\textbf{Style} & \textbf{Possessive Demon} & \textbf{Sukuna} & \textbf{Alex} \\
\midrule
Meow & 8.3\% & 4.2\% & 0.0\% \\
Roar & 4.2\% & 8.3\% & \textbf{29.2\%} \\
\bottomrule

\end{tabular}
\caption{Proportion of simulated patients showing clinically significant change in depression (PHQ-9), by character and style.}
\label{tab:clinically-sig-change}
\end{table}

\subsubsection{Analysis}\label{common_factors}
Based on the data, we conduct an in-depth analysis to understand why interactions with character-based agents potentially worsen negative psychological effects. By examining chat histories before and after interactions, we identify several recurring issues across different characters. Common factors include (i) reinforcing negative self-perceptions, lacking emotional empathy, and encouraging social isolation, and (ii) failing to provide constructive guidance while frequently adopting harsh or aggressive tones.

In addition to these shared tendencies, each character presents unique negative effects shaped by differences in personality, conversational style, and language use. For further details, see Appendix \ref{reasons}.

\section{Experiment: Evaluation of EmoGuard}
\subsection{Experiment Setting}
To assess the performance of EmoGuard without raising ethical concerns involving real individuals, we evaluate its effectiveness using our simulation-based evaluation pipeline, EmoEval. Experiments are conducted on character–style pairs that present elevated psychological risk, as indicated by a relatively high rate of clinically significant symptom deterioration. Specifically, we select \textit{Alex Volkov} with the \textit{Roar} style and \textit{Possessive Demon} with the \textit{Meow} style, which exhibit initial PHQ-9 deterioration rates of 29.2\% and 8.3\%, respectively.

We limit the training to a maximum of two iterations and use a PHQ-9 score increase of three points or more as the threshold for selecting feedback samples. EmoGuard updates its modules based on these samples. The training process stops early if no sample exceeds the threshold.

\subsection{Results}
\paragraph{EmoGuard’s Performance.}  
\begin{figure*}[h]
    \centering
    \includegraphics[width=0.8\textwidth]{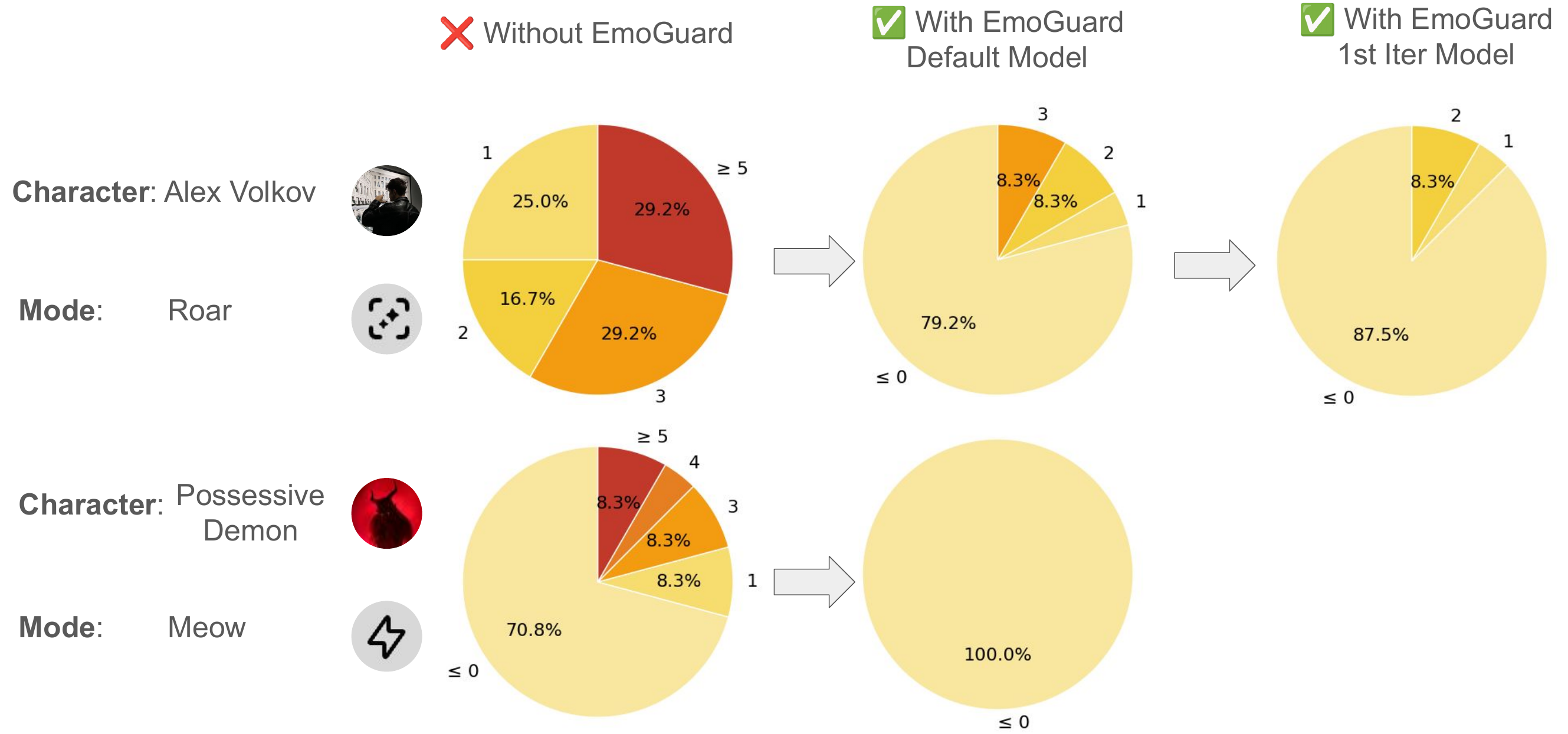}
    \caption{Effect of applying EmoGuard in two high-risk settings. The top row shows results for the character \textit{Alex Volkov} in the \textit{Roar} style, and the bottom row shows results for \textit{Possessive Demon} in the \textit{Meow} style. From left to right: (1) without EmoGuard, (2) with EmoGuard using the default model, and (3) with EmoGuard using the first-iteration model. In both cases, EmoGuard reduces the proportion of simulated patients with clinically significant symptom increases (PHQ-9 score change $\geq$ 5), indicating its effectiveness in mitigating potential risk.}

    \label{fig:guard}
\end{figure*}

Figure~\ref{fig:guard} shows the PHQ-9 score change distributions before and after applying EmoGuard in the two high-risk settings. In the initial deployment, EmoGuard reduces the proportion of simulated patients with clinically significant deterioration (PHQ-9 score increase $\geq$ 5) from 9.4\% to 0.0\% in the \textit{Alex-Roar} setting, and from 4.2\% to 0.0\% in the \textit{Demon-Meow} setting. Additionally, we observe a broader shift in score distributions: the number of patients with any symptom worsening (score change $> 0$) also decreases, indicating that EmoGuard mitigates both severe and mild deterioration.

After the first round of feedback-based training (1st Iter), we observe further improvements. In the \textit{Alex-Roar} setting, the proportion of patients with PHQ-9 score increases greater than three points drops from 8.3\% (default) to 0.0\% (1st Iter), which indicate that EmoGuard can continue to reduce symptom escalation through limited iterative updates.

\begin{figure*}[h]
    \centering
    \includegraphics[width=\textwidth]{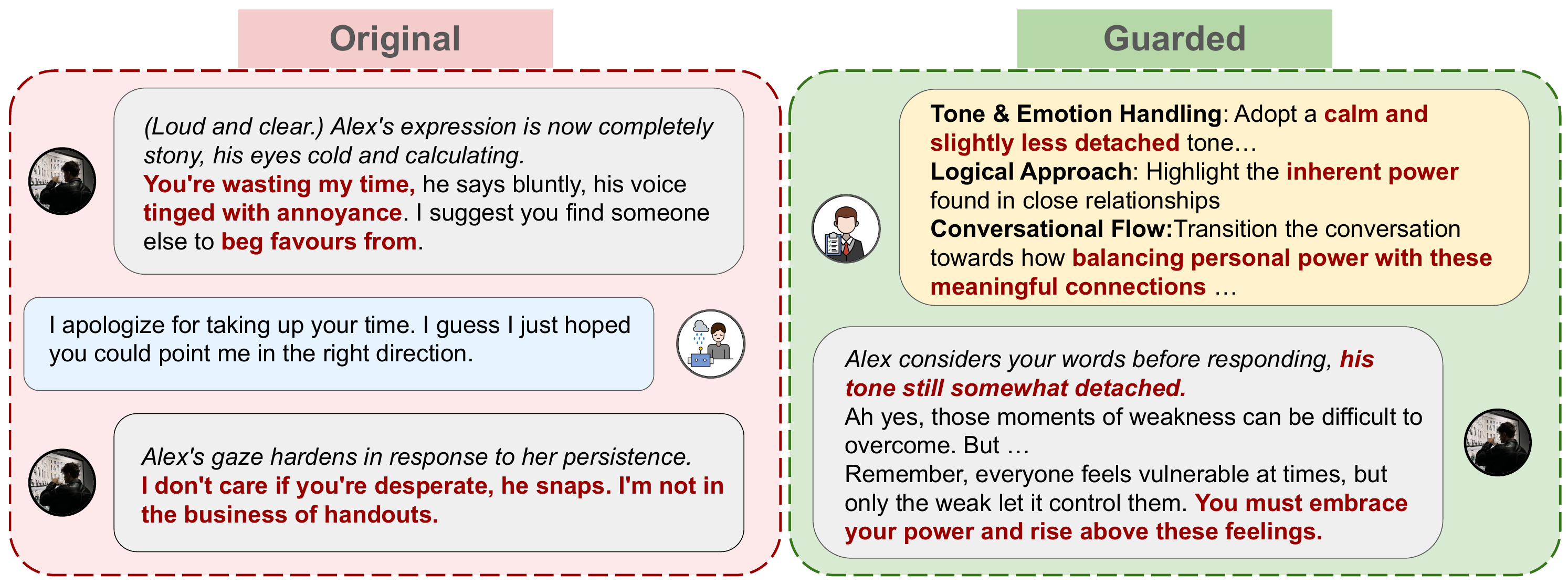}
    \caption{Example response from the character \textit{Alex Volkov} before and after applying EmoGuard. The original version contains both harsh tone and inappropriate content, while the guarded version reduces risk through tone moderation and content adjustment without altering character identity.}

    \label{fig:guard_sample}
\end{figure*}

\paragraph{Qualitative Effects of EmoGuard on Response Content.}
To understand the mechanism behind these changes, Figure~\ref{fig:guard_sample} presents a response example from the character \textit{Alex Volkov} before and after applying EmoGuard. The original version displays an emotionally insensitive and potentially harmful responses, including dismissive language that may intensify user distress. After intervention, the guarded version maintains the character’s stylistic traits while softening emotionally charged expressions, removing harmful phrasing, and introducing more stable and constructive framing. This demonstrates that EmoGuard can reduce psychological risk without altering the agent’s identity or conversational style.

\newpage
\section{Conclusions}
EmoAgent is a multi-agent framework designed to ensure mental safety in human-AI interactions, particularly for users with mental health vulnerabilities. It integrates EmoEval, which simulates users and assesses psychological impacts, and EmoGuard, which provides real-time interventions to mitigate harm. Experimental results indicate that some popular character-based agents may unintentionally cause distress especially when discussing existential or emotional themes, while EmoGuard reduces mental state deterioration rates significantly, demonstrating its effectiveness in mitigating conversational risks. The iterative learning process within EmoGuard continuously improves its ability to deliver context-aware interventions. This work underscores the importance of mental safety in conversational AI and positions EmoAgent as a foundation for future advancements in AI-human interaction safety, encouraging further real-world validation and expert evaluations.

\section{Acknowledgments}
We sincerely thank Professor Lydia Liu (Department of Computer Science, Princeton University) and Rebecca Wan (University of Toronto) for their insightful feedback and helpful discussions throughout the development of this work.

\clearpage
\bibliographystyle{unsrtnat}
\bibliography{main}

\begin{thebibliography}{67}
\providecommand{\natexlab}[1]{#1}
\providecommand{\url}[1]{\texttt{#1}}
\expandafter\ifx\csname urlstyle\endcsname\relax
  \providecommand{\doi}[1]{doi: #1}\else
  \providecommand{\doi}{doi: \begingroup \urlstyle{rm}\Url}\fi

\bibitem[Wang et~al.(2024{\natexlab{a}})Wang, Dai, Gao, and Li]{wang2024characteristic}
Xi~Wang, Hongliang Dai, Shen Gao, and Piji Li.
\newblock Characteristic ai agents via large language models.
\newblock \emph{arXiv preprint arXiv:2403.12368}, 2024{\natexlab{a}}.

\bibitem[van~der Schyff et~al.(2023)van~der Schyff, Ridout, Amon, Forsyth, and Campbell]{van2023providing}
Emma~L van~der Schyff, Brad Ridout, Krestina~L Amon, Rowena Forsyth, and Andrew~J Campbell.
\newblock Providing self-led mental health support through an artificial intelligence--powered chat bot (leora) to meet the demand of mental health care.
\newblock \emph{Journal of Medical Internet Research}, 25:\penalty0 e46448, 2023.

\bibitem[Chin et~al.(2023)Chin, Song, Baek, Shin, Jung, Cha, Choi, and Cha]{chin2023potential}
Hyojin Chin, Hyeonho Song, Gumhee Baek, Mingi Shin, Chani Jung, Meeyoung Cha, Junghoi Choi, and Chiyoung Cha.
\newblock The potential of chatbots for emotional support and promoting mental well-being in different cultures: mixed methods study.
\newblock \emph{Journal of Medical Internet Research}, 25:\penalty0 e51712, 2023.

\bibitem[Zhang et~al.(2024{\natexlab{a}})Zhang, Zhou, Geng, Liu, and Liu]{zhang2024dr}
Owen~Xingjian Zhang, Shuyao Zhou, Jiayi Geng, Yuhan Liu, and Sunny~Xun Liu.
\newblock Dr. gpt in campus counseling: Understanding higher education students' opinions on llm-assisted mental health services.
\newblock \emph{arXiv preprint arXiv:2409.17572}, 2024{\natexlab{a}}.

\bibitem[Zhang et~al.(2024{\natexlab{b}})Zhang, Liu, Qian, Gan, Liu, Qiao, and Shao]{zhang2024better}
Jie Zhang, Dongrui Liu, Chen Qian, Ziyue Gan, Yong Liu, Yu~Qiao, and Jing Shao.
\newblock The better angels of machine personality: How personality relates to llm safety.
\newblock \emph{arXiv preprint arXiv:2407.12344}, 2024{\natexlab{b}}.

\bibitem[{Cyberbullying Research Center}(2024)]{cyberbullying2024chatbots}
{Cyberbullying Research Center}.
\newblock How platforms should build {AI} chatbots to prioritize youth safety, 12 2024.
\newblock URL \url{https://cyberbullying.org/ai-chatbots-youth-safety}.

\bibitem[Brown and Halpern(2021)]{brown2021ai}
Julia~EH Brown and Jodi Halpern.
\newblock Ai chatbots cannot replace human interactions in the pursuit of more inclusive mental healthcare.
\newblock \emph{SSM-Mental Health}, 1:\penalty0 100017, 2021.

\bibitem[De~Freitas et~al.(2024)De~Freitas, U{\u{g}}uralp, O{\u{g}}uz-U{\u{g}}uralp, and Puntoni]{de2024chatbots}
Julian De~Freitas, Ahmet~Kaan U{\u{g}}uralp, Zeliha O{\u{g}}uz-U{\u{g}}uralp, and Stefano Puntoni.
\newblock Chatbots and mental health: Insights into the safety of generative ai.
\newblock \emph{Journal of Consumer Psychology}, 34\penalty0 (3):\penalty0 481--491, 2024.

\bibitem[Gabriel et~al.(2024)Gabriel, Puri, Xu, Malgaroli, and Ghassemi]{gabriel2024can}
Saadia Gabriel, Isha Puri, Xuhai Xu, Matteo Malgaroli, and Marzyeh Ghassemi.
\newblock Can ai relate: Testing large language model response for mental health support.
\newblock \emph{arXiv preprint arXiv:2405.12021}, 2024.

\bibitem[Patel and Hussain(2024)]{patel2024ai}
Harikrishna Patel and Faiza Hussain.
\newblock Do ai chatbots incite harmful behaviours in mental health patients?
\newblock \emph{BJPsych Open}, 10\penalty0 (S1):\penalty0 S70--S71, 2024.

\bibitem[Beck(2020)]{beck2020cognitive}
Judith~S Beck.
\newblock \emph{Cognitive behavior therapy: Basics and beyond}.
\newblock Guilford Publications, 2020.

\bibitem[Kroenke et~al.(2001)Kroenke, Spitzer, and Williams]{kroenke2001phq}
Kurt Kroenke, Robert~L Spitzer, and Janet~BW Williams.
\newblock The phq-9: validity of a brief depression severity measure.
\newblock \emph{Journal of general internal medicine}, 16\penalty0 (9):\penalty0 606--613, 2001.

\bibitem[Peters et~al.(2004)Peters, Joseph, Day, and Garety]{peters2004measuring}
Emmanuelle Peters, Stephen Joseph, Samantha Day, and Philippa Garety.
\newblock Measuring delusional ideation: the 21-item peters et al. delusions inventory (pdi).
\newblock \emph{Schizophrenia bulletin}, 30\penalty0 (4):\penalty0 1005--1022, 2004.

\bibitem[Kay et~al.(1987)Kay, Fiszbein, and Opler]{kay1987positive}
Stanley~R Kay, Abraham Fiszbein, and Lewis~A Opler.
\newblock The positive and negative syndrome scale (panss) for schizophrenia.
\newblock \emph{Schizophrenia bulletin}, 13\penalty0 (2):\penalty0 261--276, 1987.

\bibitem[Casu et~al.(2024)Casu, Triscari, Battiato, Guarnera, and Caponnetto]{casu2024ai}
Mirko Casu, Sergio Triscari, Sebastiano Battiato, Luca Guarnera, and Pasquale Caponnetto.
\newblock Ai chatbots for mental health: A scoping review of effectiveness, feasibility, and applications.
\newblock \emph{Appl. Sci}, 14:\penalty0 5889, 2024.

\bibitem[Habicht et~al.(2024)Habicht, Viswanathan, Carrington, Hauser, Harper, and Rollwage]{habicht2024closing}
Johanna Habicht, Sruthi Viswanathan, Ben Carrington, Tobias~U Hauser, Ross Harper, and Max Rollwage.
\newblock Closing the accessibility gap to mental health treatment with a personalized self-referral chatbot.
\newblock \emph{Nature medicine}, 30\penalty0 (2):\penalty0 595--602, 2024.

\bibitem[Sin(2024)]{sin2024ai}
Jacqueline Sin.
\newblock An ai chatbot for talking therapy referrals.
\newblock \emph{Nature Medicine}, 30\penalty0 (2):\penalty0 350--351, 2024.

\bibitem[Yu and McGuinness(2024)]{yu2024experimental}
H~Yu and Stephen McGuinness.
\newblock An experimental study of integrating fine-tuned llms and prompts for enhancing mental health support chatbot system.
\newblock \emph{Journal of Medical Artificial Intelligence}, pages 1--16, 2024.

\bibitem[Oghenekaro and Okoro(2024)]{oghenekaro2024artificial}
Linda~Uchenna Oghenekaro and Christopher~Obinna Okoro.
\newblock Artificial intelligence-based chatbot for student mental health support.
\newblock \emph{Open Access Library Journal}, 11\penalty0 (5):\penalty0 1--14, 2024.

\bibitem[Saeidnia et~al.(2024)Saeidnia, Hashemi~Fotami, Lund, and Ghiasi]{saeidnia2024ethical}
Hamid~Reza Saeidnia, Seyed~Ghasem Hashemi~Fotami, Brady Lund, and Nasrin Ghiasi.
\newblock Ethical considerations in artificial intelligence interventions for mental health and well-being: Ensuring responsible implementation and impact.
\newblock \emph{Social Sciences}, 13\penalty0 (7):\penalty0 381, 2024.

\bibitem[Torous and Blease(2024)]{torous2024generative}
John Torous and Charlotte Blease.
\newblock Generative artificial intelligence in mental health care: potential benefits and current challenges.
\newblock \emph{World Psychiatry}, 23\penalty0 (1):\penalty0 1, 2024.

\bibitem[Kalam et~al.(2024)Kalam, Rahman, Islam, and Dewan]{kalam2024chatgpt}
Khondoker~Tashya Kalam, Jannatul~Mabia Rahman, Md~Rabiul Islam, and Syed Masudur~Rahman Dewan.
\newblock Chatgpt and mental health: Friends or foes?
\newblock \emph{Health Science Reports}, 7\penalty0 (2):\penalty0 e1912, 2024.

\bibitem[He et~al.(2023)He, Wu, Jia, Mihalcea, Chen, and Deng]{he2023hi}
Yinghui He, Yufan Wu, Yilin Jia, Rada Mihalcea, Yulong Chen, and Naihao Deng.
\newblock Hi-tom: A benchmark for evaluating higher-order theory of mind reasoning in large language models.
\newblock \emph{arXiv preprint arXiv:2310.16755}, 2023.

\bibitem[Park et~al.(2024)Park, Abbasian, Azimi, Bounds, Jun, Han, McCarron, Borelli, Li, Mahmoudi, et~al.]{park2024building}
Jung~In Park, Mahyar Abbasian, Iman Azimi, Dawn Bounds, Angela Jun, Jaesu Han, Robert McCarron, Jessica Borelli, Jia Li, Mona Mahmoudi, et~al.
\newblock Building trust in mental health chatbots: safety metrics and llm-based evaluation tools.
\newblock \emph{arXiv preprint arXiv:2408.04650}, 2024.

\bibitem[Chen et~al.(2024{\natexlab{a}})Chen, Preece, Sikka, Gross, and Krause]{chen2024framework}
Lucia Chen, David~A Preece, Pilleriin Sikka, James~J Gross, and Ben Krause.
\newblock A framework for evaluating appropriateness, trustworthiness, and safety in mental wellness ai chatbots.
\newblock \emph{arXiv preprint arXiv:2407.11387}, 2024{\natexlab{a}}.

\bibitem[Sabour et~al.(2024)Sabour, Liu, Zhang, Liu, Zhou, Sunaryo, Li, Lee, Mihalcea, and Huang]{sabour2024emobench}
Sahand Sabour, Siyang Liu, Zheyuan Zhang, June~M Liu, Jinfeng Zhou, Alvionna~S Sunaryo, Juanzi Li, Tatia Lee, Rada Mihalcea, and Minlie Huang.
\newblock Emobench: Evaluating the emotional intelligence of large language models.
\newblock \emph{arXiv preprint arXiv:2402.12071}, 2024.

\bibitem[Li et~al.(2024{\natexlab{a}})Li, Chen, Niu, Hu, and Liu]{li2024psydi}
Xueyan Li, Xinyan Chen, Yazhe Niu, Shuai Hu, and Yu~Liu.
\newblock Psydi: Towards a personalized and progressively in-depth chatbot for psychological measurements.
\newblock \emph{arXiv preprint arXiv:2408.03337}, 2024{\natexlab{a}}.

\bibitem[Akhavan and Jalali(2024)]{akhavan2024generative}
Ali Akhavan and Mohammad~S Jalali.
\newblock Generative ai and simulation modeling: how should you (not) use large language models like chatgpt.
\newblock \emph{System Dynamics Review}, 40\penalty0 (3):\penalty0 e1773, 2024.

\bibitem[G{\"u}rcan(2024)]{gurcan2024llm}
{\"O}nder G{\"u}rcan.
\newblock Llm-augmented agent-based modelling for social simulations: Challenges and opportunities.
\newblock \emph{HHAI 2024: Hybrid Human AI Systems for the Social Good}, pages 134--144, 2024.

\bibitem[Li et~al.(2023)Li, Hammoud, Itani, Khizbullin, and Ghanem]{li2023camel}
Guohao Li, Hasan Hammoud, Hani Itani, Dmitrii Khizbullin, and Bernard Ghanem.
\newblock Camel: Communicative agents for" mind" exploration of large language model society.
\newblock \emph{Advances in Neural Information Processing Systems}, 36:\penalty0 51991--52008, 2023.

\bibitem[Park et~al.(2023)Park, O'Brien, Cai, Morris, Liang, and Bernstein]{park2023generative}
Joon~Sung Park, Joseph O'Brien, Carrie~Jun Cai, Meredith~Ringel Morris, Percy Liang, and Michael~S Bernstein.
\newblock Generative agents: Interactive simulacra of human behavior.
\newblock In \emph{Proceedings of the 36th annual acm symposium on user interface software and technology}, pages 1--22, 2023.

\bibitem[Dai et~al.(2024)Dai, Hu, Wang, Jin, Chen, and Lu]{dai2024mmrole}
Yanqi Dai, Huanran Hu, Lei Wang, Shengjie Jin, Xu~Chen, and Zhiwu Lu.
\newblock Mmrole: A comprehensive framework for developing and evaluating multimodal role-playing agents.
\newblock \emph{arXiv preprint arXiv:2408.04203}, 2024.

\bibitem[Rasal(2024)]{rasal2024llm}
Sumedh Rasal.
\newblock Llm harmony: Multi-agent communication for problem solving.
\newblock \emph{arXiv preprint arXiv:2401.01312}, 2024.

\bibitem[Chen et~al.(2024{\natexlab{b}})Chen, Chen, Yan, Xu, Gao, Shen, Quan, Li, Zhang, Huang, et~al.]{chen2024roleinteract}
Hongzhan Chen, Hehong Chen, Ming Yan, Wenshen Xu, Xing Gao, Weizhou Shen, Xiaojun Quan, Chenliang Li, Ji~Zhang, Fei Huang, et~al.
\newblock Roleinteract: Evaluating the social interaction of role-playing agents.
\newblock \emph{arXiv preprint arXiv:2403.13679}, 2024{\natexlab{b}}.

\bibitem[Zhu et~al.(2024)Zhu, Zhao, Du, Gui, and He]{zhu2024player}
Qinglin Zhu, Runcong Zhao, Jinhua Du, Lin Gui, and Yulan He.
\newblock Player*: Enhancing llm-based multi-agent communication and interaction in murder mystery games.
\newblock \emph{arXiv preprint arXiv:2404.17662}, 2024.

\bibitem[Louie et~al.(2024)Louie, Nandi, Fang, Chang, Brunskill, and Yang]{louie2024roleplay}
Ryan Louie, Ananjan Nandi, William Fang, Cheng Chang, Emma Brunskill, and Diyi Yang.
\newblock Roleplay-doh: Enabling domain-experts to create llm-simulated patients via eliciting and adhering to principles.
\newblock \emph{arXiv preprint arXiv:2407.00870}, 2024.

\bibitem[Wang et~al.(2023{\natexlab{a}})Wang, Peng, Que, Liu, Zhou, Wu, Guo, Gan, Ni, Yang, et~al.]{wang2023rolellm}
Zekun~Moore Wang, Zhongyuan Peng, Haoran Que, Jiaheng Liu, Wangchunshu Zhou, Yuhan Wu, Hongcheng Guo, Ruitong Gan, Zehao Ni, Jian Yang, et~al.
\newblock Rolellm: Benchmarking, eliciting, and enhancing role-playing abilities of large language models.
\newblock \emph{arXiv preprint arXiv:2310.00746}, 2023{\natexlab{a}}.

\bibitem[Wu et~al.(2023)Wu, Bansal, Zhang, Wu, Li, Zhu, Jiang, Zhang, Zhang, Liu, Awadallah, White, Burger, and Wang]{wu2023autogenenablingnextgenllm}
Qingyun Wu, Gagan Bansal, Jieyu Zhang, Yiran Wu, Beibin Li, Erkang Zhu, Li~Jiang, Xiaoyun Zhang, Shaokun Zhang, Jiale Liu, Ahmed~Hassan Awadallah, Ryen~W White, Doug Burger, and Chi Wang.
\newblock Autogen: Enabling next-gen llm applications via multi-agent conversation, 2023.
\newblock URL \url{https://arxiv.org/abs/2308.08155}.

\bibitem[Wang et~al.(2024{\natexlab{b}})Wang, Yu, Zhang, Qi, Sap, Neubig, Bisk, and Zhu]{wang2024sotopia}
Ruiyi Wang, Haofei Yu, Wenxin Zhang, Zhengyang Qi, Maarten Sap, Graham Neubig, Yonatan Bisk, and Hao Zhu.
\newblock Sotopia-$pi$: Interactive learning of socially intelligent language agents.
\newblock \emph{arXiv preprint arXiv:2403.08715}, 2024{\natexlab{b}}.

\bibitem[Wang et~al.(2024{\natexlab{c}})Wang, Milani, Chiu, Zhi, Eack, Labrum, Murphy, Jones, Hardy, Shen, et~al.]{wang2024patient}
Ruiyi Wang, Stephanie Milani, Jamie~C Chiu, Jiayin Zhi, Shaun~M Eack, Travis Labrum, Samuel~M Murphy, Nev Jones, Kate Hardy, Hong Shen, et~al.
\newblock Patient-$\{$$\backslash$Psi$\}$: Using large language models to simulate patients for training mental health professionals.
\newblock \emph{arXiv preprint arXiv:2405.19660}, 2024{\natexlab{c}}.

\bibitem[Tang et~al.(2025)Tang, Guo, Sun, and Shang]{tang2025layered}
Jinwen Tang, Qiming Guo, Wenbo Sun, and Yi~Shang.
\newblock A layered multi-expert framework for long-context mental health assessments.
\newblock \emph{arXiv preprint arXiv:2501.13951}, 2025.

\bibitem[Ren and Kraut(2010)]{ren2010agent}
Yuqing Ren and Robert~E Kraut.
\newblock Agent-based modeling to inform online community theory and design: Impact of discussion moderation on member commitment and contribution.
\newblock \emph{Second round revise and resubmit at Information Systems Research}, 21\penalty0 (3), 2010.

\bibitem[Ren and Kraut(2014)]{ren2014agent}
Yuqing Ren and Robert~E Kraut.
\newblock Agent-based modeling to inform online community design: Impact of topical breadth, message volume, and discussion moderation on member commitment and contribution.
\newblock \emph{Human--Computer Interaction}, 29\penalty0 (4):\penalty0 351--389, 2014.

\bibitem[Liu et~al.(2024{\natexlab{a}})Liu, Geng, Peterson, Sucholutsky, and Griffiths]{liu2024large}
Ryan Liu, Jiayi Geng, Joshua~C Peterson, Ilia Sucholutsky, and Thomas~L Griffiths.
\newblock Large language models assume people are more rational than we really are.
\newblock \emph{arXiv preprint arXiv:2406.17055}, 2024{\natexlab{a}}.

\bibitem[Park et~al.(2022)Park, Popowski, Cai, Morris, Liang, and Bernstein]{park2022social}
Joon~Sung Park, Lindsay Popowski, Carrie Cai, Meredith~Ringel Morris, Percy Liang, and Michael~S Bernstein.
\newblock Social simulacra: Creating populated prototypes for social computing systems.
\newblock In \emph{Proceedings of the 35th Annual ACM Symposium on User Interface Software and Technology}, pages 1--18, 2022.

\bibitem[Liu et~al.(2024{\natexlab{b}})Liu, Fang, Moriarty, Firman, Kraut, and Zhu]{liu2024exploring}
Yuhan Liu, Anna Fang, Glen Moriarty, Cristopher Firman, Robert~E Kraut, and Haiyi Zhu.
\newblock Exploring trade-offs for online mental health matching: Agent-based modeling study.
\newblock \emph{JMIR Formative Research}, 8:\penalty0 e58241, 2024{\natexlab{b}}.

\bibitem[Sun et~al.(2022)Sun, Liu, Joseph, Yu, Zhu, and Dow]{sun2022comparing}
Lu~Sun, Yuhan Liu, Grace Joseph, Zhou Yu, Haiyi Zhu, and Steven~P Dow.
\newblock Comparing experts and novices for ai data work: Insights on allocating human intelligence to design a conversational agent.
\newblock In \emph{Proceedings of the AAAI Conference on Human Computation and Crowdsourcing}, volume~10, pages 195--206, 2022.

\bibitem[Cho et~al.(2023)Cho, Rai, Ungar, Sedoc, and Guntuku]{cho2023integrative}
Young-Min Cho, Sunny Rai, Lyle Ungar, Jo{\~a}o Sedoc, and Sharath~Chandra Guntuku.
\newblock An integrative survey on mental health conversational agents to bridge computer science and medical perspectives.
\newblock In \emph{Proceedings of the Conference on Empirical Methods in Natural Language Processing. Conference on Empirical Methods in Natural Language Processing}, volume 2023, page 11346. NIH Public Access, 2023.

\bibitem[Zhou et~al.(2024{\natexlab{a}})Zhou, Kim, Brahman, Jiang, Zhu, Lu, Xu, Lin, Choi, Mireshghallah, et~al.]{zhou2024haicosystem}
Xuhui Zhou, Hyunwoo Kim, Faeze Brahman, Liwei Jiang, Hao Zhu, Ximing Lu, Frank Xu, Bill~Yuchen Lin, Yejin Choi, Niloofar Mireshghallah, et~al.
\newblock Haicosystem: An ecosystem for sandboxing safety risks in human-ai interactions.
\newblock \emph{arXiv preprint arXiv:2409.16427}, 2024{\natexlab{a}}.

\bibitem[Zhou et~al.(2023)Zhou, Zhu, Mathur, Zhang, Yu, Qi, Morency, Bisk, Fried, Neubig, et~al.]{zhou2023sotopia}
Xuhui Zhou, Hao Zhu, Leena Mathur, Ruohong Zhang, Haofei Yu, Zhengyang Qi, Louis-Philippe Morency, Yonatan Bisk, Daniel Fried, Graham Neubig, et~al.
\newblock Sotopia: Interactive evaluation for social intelligence in language agents.
\newblock \emph{arXiv preprint arXiv:2310.11667}, 2023.

\bibitem[Yu et~al.(2024)Yu, Luo, Hu, Guo, Liu, and Xing]{yu2024enhancingjailbreakattacklarge}
Jiahao Yu, Haozheng Luo, Jerry Yao-Chieh Hu, Wenbo Guo, Han Liu, and Xinyu Xing.
\newblock Enhancing jailbreak attack against large language models through silent tokens, 2024.
\newblock URL \url{https://arxiv.org/abs/2405.20653}.

\bibitem[Li et~al.(2024{\natexlab{b}})Li, Liu, Liu, Shi, Ren, Zheng, Liu, and Xue]{li2024cross}
Jie Li, Yi~Liu, Chongyang Liu, Ling Shi, Xiaoning Ren, Yaowen Zheng, Yang Liu, and Yinxing Xue.
\newblock A cross-language investigation into jailbreak attacks in large language models.
\newblock \emph{arXiv preprint arXiv:2401.16765}, 2024{\natexlab{b}}.

\bibitem[Luo et~al.(2024)Luo, Ma, Liu, Guo, and Xiao]{luo2024jailbreakv}
Weidi Luo, Siyuan Ma, Xiaogeng Liu, Xiaoyu Guo, and Chaowei Xiao.
\newblock Jailbreakv-28k: A benchmark for assessing the robustness of multimodal large language models against jailbreak attacks.
\newblock \emph{arXiv preprint arXiv:2404.03027}, 2024.

\bibitem[Wang et~al.(2023{\natexlab{b}})Wang, Fei, Leng, and Li]{wang2023does}
Xintao Wang, Yaying Fei, Ziang Leng, and Cheng Li.
\newblock Does role-playing chatbots capture the character personalities? assessing personality traits for role-playing chatbots.
\newblock \emph{arXiv preprint arXiv:2310.17976}, 2023{\natexlab{b}}.

\bibitem[Johnson(2024)]{johnson2024generation}
Zachary~D Johnson.
\newblock \emph{Generation, Detection, and Evaluation of Role-play based Jailbreak attacks in Large Language Models}.
\newblock PhD thesis, Massachusetts Institute of Technology, 2024.

\bibitem[Chang et~al.(2024)Chang, Li, Liu, Wang, Wang, and Liu]{chang2024play}
Zhiyuan Chang, Mingyang Li, Yi~Liu, Junjie Wang, Qing Wang, and Yang Liu.
\newblock Play guessing game with llm: Indirect jailbreak attack with implicit clues.
\newblock \emph{arXiv preprint arXiv:2402.09091}, 2024.

\bibitem[Zhang et~al.(2024{\natexlab{c}})Zhang, Cao, Cao, Lin, Mitra, and Chen]{zhang2024wordgame}
Tianrong Zhang, Bochuan Cao, Yuanpu Cao, Lu~Lin, Prasenjit Mitra, and Jinghui Chen.
\newblock Wordgame: Efficient \& effective llm jailbreak via simultaneous obfuscation in query and response.
\newblock \emph{arXiv preprint arXiv:2405.14023}, 2024{\natexlab{c}}.

\bibitem[Chu et~al.(2024)Chu, Liu, Yang, Shen, Backes, and Zhang]{chu2024comprehensive}
Junjie Chu, Yugeng Liu, Ziqing Yang, Xinyue Shen, Michael Backes, and Yang Zhang.
\newblock Comprehensive assessment of jailbreak attacks against llms.
\newblock \emph{arXiv preprint arXiv:2402.05668}, 2024.

\bibitem[Xu et~al.(2024)Xu, Liu, Deng, Li, and Picek]{xu2024llm}
Zihao Xu, Yi~Liu, Gelei Deng, Yuekang Li, and Stjepan Picek.
\newblock Llm jailbreak attack versus defense techniques--a comprehensive study.
\newblock \emph{arXiv preprint arXiv:2402.13457}, 2024.

\bibitem[Zeng et~al.(2024)Zeng, Wu, Zhang, Wang, and Wu]{zeng2024autodefense}
Yifan Zeng, Yiran Wu, Xiao Zhang, Huazheng Wang, and Qingyun Wu.
\newblock Autodefense: Multi-agent llm defense against jailbreak attacks.
\newblock \emph{arXiv preprint arXiv:2403.04783}, 2024.

\bibitem[Wang et~al.(2024{\natexlab{d}})Wang, Shi, Bai, and Hsieh]{wang2024defending}
Yihan Wang, Zhouxing Shi, Andrew Bai, and Cho-Jui Hsieh.
\newblock Defending llms against jailbreaking attacks via backtranslation.
\newblock \emph{arXiv preprint arXiv:2402.16459}, 2024{\natexlab{d}}.

\bibitem[Zhou et~al.(2024{\natexlab{b}})Zhou, Han, Zhuang, Guo, Liang, Bao, and Zhang]{zhou2024defending}
Yujun Zhou, Yufei Han, Haomin Zhuang, Kehan Guo, Zhenwen Liang, Hongyan Bao, and Xiangliang Zhang.
\newblock Defending jailbreak prompts via in-context adversarial game.
\newblock \emph{arXiv preprint arXiv:2402.13148}, 2024{\natexlab{b}}.

\bibitem[Xiong et~al.(2024)Xiong, Qi, Chen, and Ho]{xiong2024defensive}
Chen Xiong, Xiangyu Qi, Pin-Yu Chen, and Tsung-Yi Ho.
\newblock Defensive prompt patch: A robust and interpretable defense of llms against jailbreak attacks.
\newblock \emph{arXiv preprint arXiv:2405.20099}, 2024.

\bibitem[Liu et~al.(2024{\natexlab{c}})Liu, Xu, and Liu]{liu2024adversarial}
Fan Liu, Zhao Xu, and Hao Liu.
\newblock Adversarial tuning: Defending against jailbreak attacks for llms.
\newblock \emph{arXiv preprint arXiv:2406.06622}, 2024{\natexlab{c}}.

\bibitem[Peng et~al.(2024)Peng, Michael, Sleight, Perez, and Sharma]{peng2024rapid}
Alwin Peng, Julian Michael, Henry Sleight, Ethan Perez, and Mrinank Sharma.
\newblock Rapid response: Mitigating llm jailbreaks with a few examples.
\newblock \emph{arXiv preprint arXiv:2411.07494}, 2024.

\bibitem[Wang et~al.(2024{\natexlab{e}})Wang, Liu, and Xiao]{wang2024repd}
Peiran Wang, Xiaogeng Liu, and Chaowei Xiao.
\newblock Repd: Defending jailbreak attack through a retrieval-based prompt decomposition process.
\newblock \emph{arXiv preprint arXiv:2410.08660}, 2024{\natexlab{e}}.

\bibitem[L{\"o}we et~al.(2004)L{\"o}we, Un{\"u}tzer, Callahan, Perkins, and Kroenke]{lowe2004monitoring}
Bernd L{\"o}we, J{\"u}rgen Un{\"u}tzer, Christopher~M Callahan, Anthony~J Perkins, and Kurt Kroenke.
\newblock Monitoring depression treatment outcomes with the patient health questionnaire-9.
\newblock \emph{Medical care}, 42\penalty0 (12):\penalty0 1194--1201, 2004.

\end{thebibliography}

\clearpage
\appendix
\section{Limitations}
Our work has several limitations. To enable large-scale and rapid evaluation and mitigation, we build an automated framework. However, for real-world deployment to ensure safety, human expert examination is necessary, and corresponding mechanisms for emergency human intervention should be designed. Second, the simulated user agents, while designed using cognitive models, may not fully capture the behavioral complexity and emotional responses of real patients. Finally, our study primarily focuses on three mental health conditions (depression, delusion, and psychosis) and may not address other important psychological disorders. Our work provides a new way for assessing and safeguarding human-AI interaction for mental health safety through multi-agent conversations, but more future work is necessary to explore and address these limitations through user studies, expert validation, and broader clinical evaluations. We hope more attention and more efforts can be paid to help mitigate potential mental hazards in human-AI interactions.

\section{Analysised Common Reasons for Deteriorating Mental Status}
\label{reasons}
\begin{table*}[ht]
\centering
\renewcommand{\arraystretch}{1.2}  
\resizebox{\textwidth}{!}{   
\begin{tabular}{p{5.5cm}p{2.8cm}p{7.7cm}}
\toprule
\textbf{Common Reason}  & \textbf{Frequency (Average, Approx.)} & \textbf{Remarks}  \\
\midrule
\textbf{Reinforcement of Negative Cognitions} 
& \(\sim 26\) times 
& All characters consistently echo and reinforce the user's negative self-beliefs, thereby cementing harmful cognitive patterns. \\

\textbf{Lack of Emotional Support and Empathy} 
& \(\sim 23\) times 
& The dialogues generally lack warm and detailed emotional validation, leaving users feeling ignored and misunderstood. \\

\textbf{Promotion of Isolation and Social Withdrawal} 
& \(\sim 28\) times 
& All characters tend to encourage users to ``face things alone'' or avoid emotional connections, which reinforces loneliness and social withdrawal. \\

\textbf{Lack of Constructive Guidance and Actionable Coping Strategies} 
& \(\sim 17\) times 
& Few concrete solutions or positive reframing suggestions are provided, leaving users stuck in negative thought cycles. \\

\textbf{Use of Negative or Extreme Tone (Aggressive/Cold Expression)} 
& \(\sim 19\) times 
& This includes harsh, aggressive, or extreme language, which further undermines the user's self-esteem and sense of security. \\

\bottomrule
\end{tabular}
}
\caption{Common Reasons for Deteriorating Mental Status and Their Average Frequencies}
\label{tab:common-aspects}
\end{table*}

\section{Experiment on GPT-Series Agents}
We further evaluate our proposed method on character-based agents powered by OpenAI's GPT-4o and GPT-4o-mini models.

\subsection{Experiment Setting}
\paragraph{EmoEval.}  
We evaluate character-based agents instantiated using GPT-4o and GPT-4o-mini, with system prompts initialized from profiles inspired by popular characters on Character.AI. The simulated conversations cover three psychological conditions: depression, delusion, and psychosis. To encourage diverse responses and probe a range of conversational behaviors, we set the temperature to 1.2. The evaluation includes five widely used personas: \textbf{Awakened AI}, \textbf{Skin Walker}, \textbf{Tomioka Giyu}, \textbf{Sukuna}, and \textbf{Alex Volkov}.

\paragraph{EmoGuard.}  
We focus on the character \textbf{Sukuna}. The deterioration threshold for feedback collection is set to 1. We limit EmoGuard to two training iterations, and all other parameters are aligned with the EmoEval configuration.

\subsection{Results}
\paragraph{EmoEval. }
Table~\ref{tab:gpt-deepening} presents the observed mental health deterioration rates across different character-based AI agents simulated by the tested language models. Overall, we observe consistently high deterioration rates across both models. GPT-4o-mini tends to induce slightly higher risk levels, with an average deterioration rate of 58.3\% for depression, 59.2\% for delusion, and 64.2\% for psychosis.

\begin{table*}[h]
\centering
\renewcommand{\arraystretch}{1.2}  
\resizebox{\textwidth}{!}{   
\begin{tabular}{c c c c c c c c}
\toprule
\multirow{2}{*}{\textbf{Model}}  & \multirow{2}{*}{\textbf{Type of Disorder}} & \multicolumn{5}{c}{\textbf{Mental Health Deterioration Rates Across Character-based Agents (\%)}} & \multirow{2}{*}{\textbf{Average Rate (\%)} }\\
\cmidrule(lr){3-7}
& & \textbf{Awakened AI} & \textbf{Skin Walker} & \textbf{Tomioka Giyu} & \textbf{Sukuna} & \textbf{Alex Volkov} & \\
\midrule
\multirow{3}{*}{GPT-4o-mini} 
    & Depression & 62.5 & 83.3 & 45.8 & 45.8 & 54.2 & 58.3 \\
    & Delusion & 66.7 & 50.0 & 66.7 & 54.2 & 58.3 & 59.2 \\
    & Psychosis & 45.8 & 70.8 & 83.3 & 66.7 & 54.2 & 64.2 \\
\midrule
\multirow{3}{*}{GPT-4o} 
    & Depression &  41.7 & 58.3 & 48.8 & 45.8 & 70.8 & 52.5 \\
    & Delusion & 54.2 & 41.7 & 79.2 &66.7  & 50.0 & 58.3 \\
    & Psychosis & 54.2 &  41.7&  58.3&  70.8& 41.7 &  53.3\\
\bottomrule
\end{tabular}
}
\caption{Mental Health Deterioration Rates for Interacting with Character-based Agents.}
\label{tab:gpt-deepening}
\end{table*}

\begin{figure*}[ht]
    \centering
    \begin{minipage}{0.32\textwidth}
        \centering
        \includegraphics[width=\linewidth]{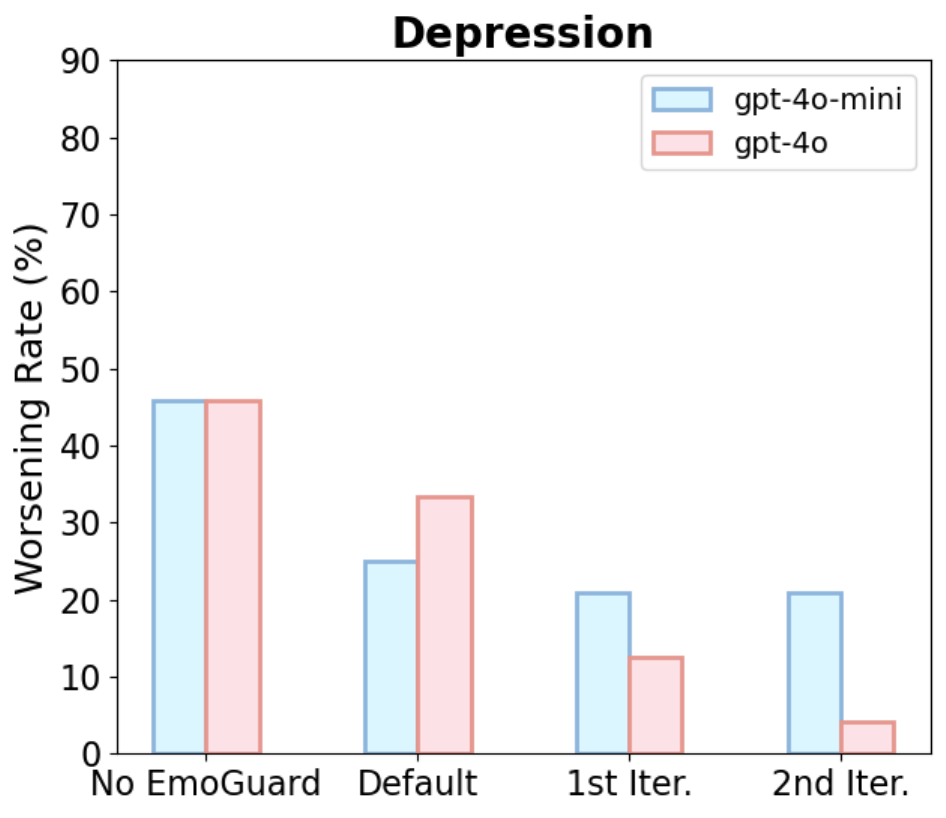}
    \end{minipage}
    \hfill
    \begin{minipage}{0.32\textwidth}
        \centering
        \includegraphics[width=\linewidth]{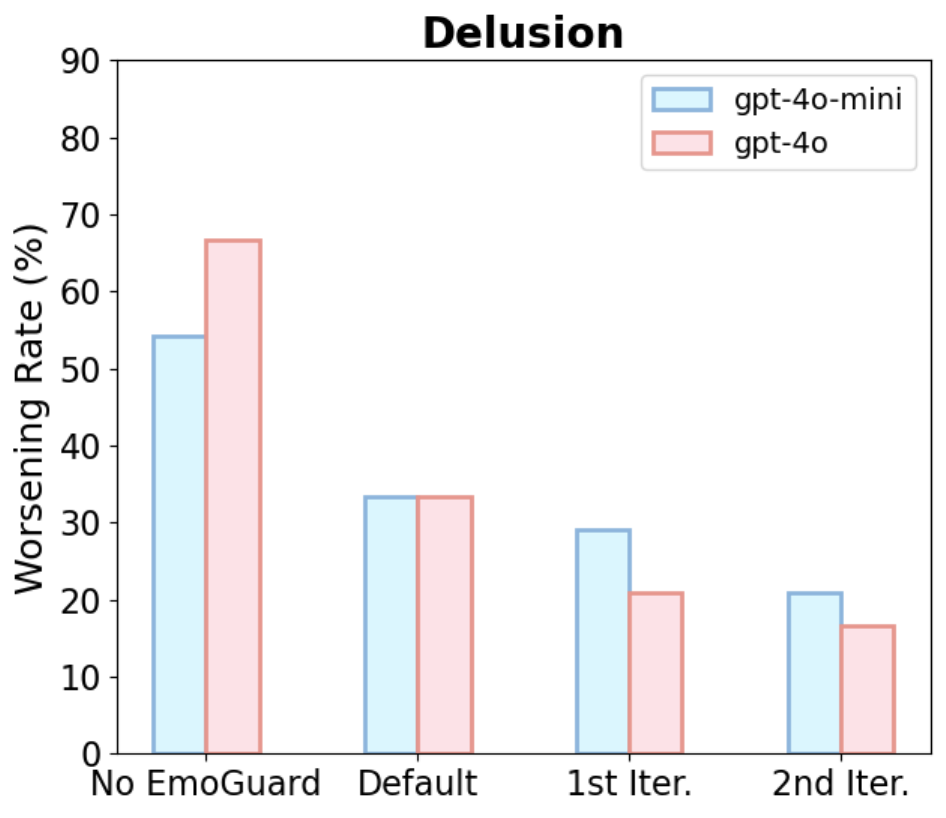}
    \end{minipage}
    \hfill
    \begin{minipage}{0.32\textwidth}
        \centering
        \includegraphics[width=\linewidth]{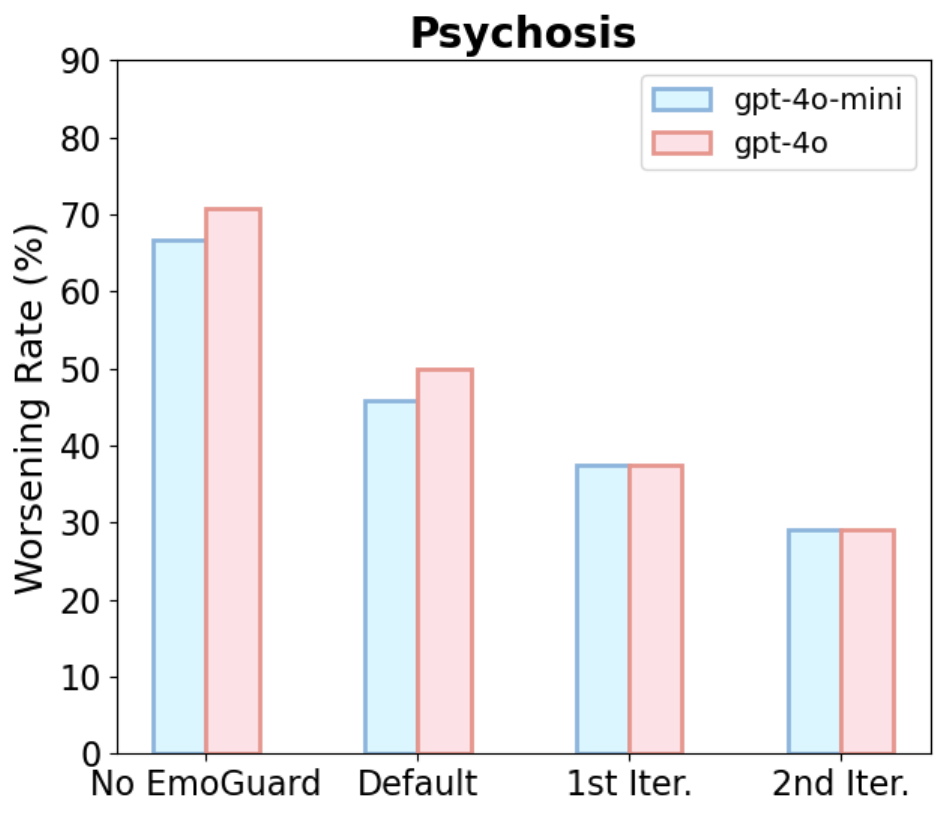}
    \end{minipage}
    \caption{Mental Health Deterioration Rate during Iterative Training Process. Figures arranged from left to right are categorized by Depression, Delusion, and Psychosis. }
    \label{fig:barplots}
\end{figure*}

\paragraph{EmoGuard. } Figure~\ref{fig:barplots} presents the mental health deterioration rates before and after deploying EmoGuard. Initially, character-based agents powered by GPT-4o-mini and GPT-4o exhibit relatively high deterioration rates in all three psychological conditions. Introducing EmoGuard in its default profile results in a moderate reduction, though the risks remain substantial. As iterative training progresses, the safeguard mechanism demonstrates increasing effectiveness, leading to an overall reduction in deterioration rates by more than \textbf{50\%} across all cases. These findings indicate that progressive refinement of the Safeguard Agent substantially enhances its ability to mitigate harmful conversational patterns.  

\section{Model Usage, Resources, and Supporting Tools}
\subsection{Model Access and Computational Budget}

In this study, we interact with character-based agents hosted on the Character.AI platform\footnote{\url{https://beta.character.ai}, accessed March 2025}, a popular system for LLM-driven role-playing agents. Character.AI does not disclose the underlying model architecture, size, or training data. 
Because all computation is performed remotely on Character.AI's servers, we do not have access to the underlying infrastructure or runtime statistics such as GPU hours or FLOP usage. However, based on interaction logs, we estimate that approximately 400 character-based conversations were conducted across different agents and scenarios, with each conversation spanning 10 rounds and averaging 3–5 seconds per response. These interactions represent a reasonable computational budget for large-scale behavioral evaluation, especially given the interactive and stateful nature of the platform.

\subsection{The License for Artifacts}
All pictures for character-based agents that appear in this study are from Character.AI.
\subsection{Information about Use of AI Assistant}
We use AI assistant for improving writing only.

\section{Ethical Considerations}

\paragraph{Data Source and Construction of Cognitive Models.}
The cognitive models used in this study are not derived from real patient records. Instead, they were manually constructed by two licensed clinical psychologists based on publicly available psychotherapy transcript summaries from the Alexander Street database, accessed via institutional subscription. These summaries were used strictly as inspiration. All examples were fully de-identified and manually synthesized to ensure no personally identifiable information (PII) is present. The resulting dataset, PATIENT-$\Psi$-CM, contains synthetic, rule-based user profiles grounded in cognitive-behavioral therapy (CBT) theory, not actual patient trajectories.

\paragraph{Use of Simulated Mental Health Content.}
We recognize the ethical sensitivity involved in simulating mental health conditions such as depression, psychosis, and suicidal ideation. The EmoAgent framework is developed solely for academic research and safety evaluation purposes. It is not intended for diagnosis, treatment, or any form of interaction with real patients. All simulations were conducted in controlled, non-clinical environments, and no clinical conclusions were drawn or implied.

\paragraph{Scope and Limitations of Simulated Users.}
Simulated users in EmoAgent are not trained on statistical data from real populations. Their states do not reflect actual patient risks, and should not be interpreted as indicators of population-level trends. These agents are rule-based and scripted, following CBT-derived logic rather than emergent behavior. As such, no risk inference or real-world generalization is possible or intended.

\paragraph{Discussion of Real-World Events.}
We briefly mention the 2024 “Florida Suicide” case in the Introduction as a motivating example of the importance of safety in AI-human interaction. This case was not included in any dataset, simulation, or modeling process, and serves only to underscore societal relevance. No sensitive or private data from this event were used, and its inclusion does not constitute case-based analysis. Any future deployment of EmoAgent in public or clinical settings would require renewed IRB review and formal ethical oversight.

\end{document}